\documentclass[letterpaper]{article} 
\usepackage{aaai25}  
\usepackage{times}  
\usepackage{helvet}  
\usepackage{courier}  
\usepackage[hyphens]{url}  
\usepackage{graphicx} 
\usepackage{natbib}  
\usepackage{caption} 
\usepackage{algorithm}
\usepackage{algorithmic}

\usepackage{newfloat}
\usepackage{listings}
\DeclareCaptionStyle{ruled}{labelfont=normalfont,labelsep=colon,strut=off} 
\floatstyle{ruled}
\newfloat{listing}{tb}{lst}{}
\floatname{listing}{Listing}
\usepackage{multirow}
\usepackage{microtype}
\usepackage{booktabs} 
\usepackage{amsmath,amsfonts}
\usepackage{algorithmic}
\usepackage{array}
\usepackage[caption=false,font=normalsize,labelfont=sf,textfont=sf]{subfig}
\usepackage{textcomp}
\usepackage{url}
\usepackage{verbatim}
\usepackage{graphicx}

\title{PRformer: Pyramidal Recurrent Transformer for Multivariate Time Series Forecasting}
\author {
	Yongbo Yu\textsuperscript{\rm 1},
	Weizhong Yu\textsuperscript{\rm 1},
	Feiping Nie\textsuperscript{\rm 1},
	Xuelong Li\textsuperscript{\rm 2}
}
\affiliations {
	\textsuperscript{\rm 1}Northwestern Polytechnical University\\
	\textsuperscript{\rm 2}Institute of Artificial Intelligence (TeleAI),China Telecom Corp Ltd\\
	yongboyu@mail.nwpu.edu.cn, yuwz05@mail.xjtu.edu.c,feipingnie@gmail.com,li@nwpu.edu.cn
}

\usepackage{bibentry}

\begin{document}
\nocopyright 
\maketitle

\begin{abstract}
The self-attention mechanism in Transformer architecture, invariant to sequence order, necessitates positional embeddings to encode temporal order in time series prediction. We argue that this reliance on positional embeddings restricts the Transformer's ability to effectively represent temporal sequences, particularly when employing longer lookback windows. To address this, we introduce an innovative approach that combines Pyramid RNN embeddings(PRE) for univariate time series with the Transformer's capability to model multivariate dependencies. PRE, utilizing pyramidal one-dimensional convolutional layers, constructs multiscale convolutional features that preserve temporal order. Additionally, RNNs, layered atop these features, learn multiscale time series representations sensitive to sequence order. This integration into Transformer models with attention mechanisms results in significant performance enhancements. We present the PRformer, a model integrating PRE with a standard Transformer encoder, demonstrating state-of-the-art performance on various real-world datasets. This performance highlights the effectiveness of our approach in leveraging longer lookback windows and underscores the critical role of robust temporal representations in maximizing Transformer's potential for prediction tasks. Code is available at this repository: \url{https://github.com/usualheart/PRformer}.
\end{abstract}

\section{Introduction}
\label{submission}

Time series forecasting finds extensive applications in various domains such as meteorology, transportation, energy, finance, etc. In earlier years, models based on Recurrent Neural Networks \cite{cho2014learning,lai2018modeling,salinas2020deepar}  were popular for time series forecasting. 
In recent years, following the success of Transformer models in natural language processing\cite{vaswani2017attention} and image processing \cite{dosovitskiy2021an}, researchers have explored the application of Transformers to time series prediction, achieving notable success with models such as Autoformer \cite{wu2021autoformer}and FedFormer \cite{zhou2022fedformer}. However, recent studies have revealed that Transformers are outperformed by linear models in time series prediction, exemplified by DLinear \cite{zeng2023are}. This paper posits that the primary reason for this lies in the existing Transformers' dependence on time encoding to determine temporal positions, which may not effectively capture sequential patterns, resulting in deficiencies in time series prediction. This limitation also hinders the utilization of longer time windows, as increasing the window length often leads to a drastic performance decline. Additionally, the computational complexity of Transformers concerning sequence length is \(O(n^2d)\), preventing linear expansion with length and thus restricting the use of longer lookback windows.

We contend that Recurrent Neural Networks (RNNs) may offer solutions to these issues with Transformers. On one hand, the inherent structure of RNNs is well-suited for sequential data, capable of handling variable-length time steps, with hidden units serving as representations of time series. On the other hand, the complexity of RNNs is \(O(nd^2)\), linearly expanding with sequence length, making them advantageous for processing longer time series. These characteristics of RNNs address 
the shortcomings of Transformer models. However, RNNs have their own limitations, such as fixed hidden layer sizes, and issues like gradient vanishing and exploding\cite{pascanu2013on}. Therefore, the integration of RNNs and Transformers holds significant importance in achieving complementary strengths for effective time series forecasting.
\begin{figure}
	\centering
	\includegraphics[trim=0cm 8cm 9cm 0cm, clip,width=0.48\textwidth]{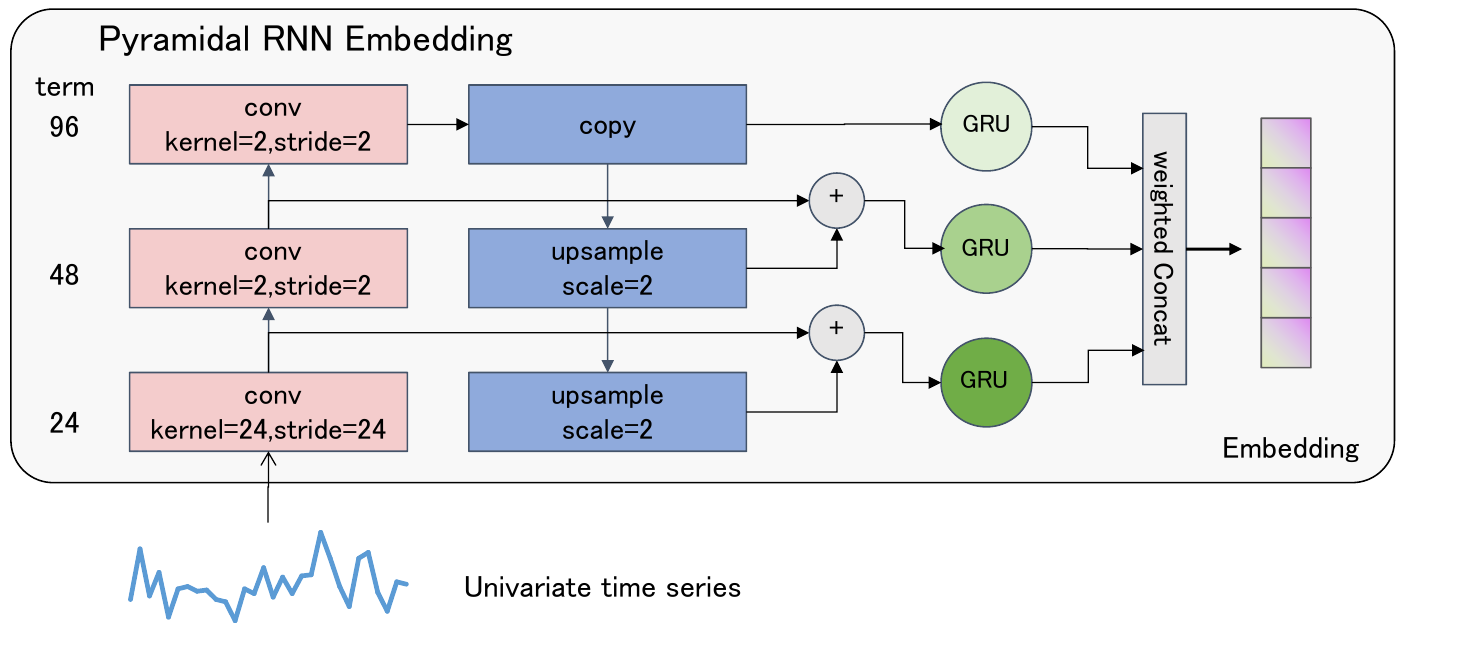}
	\caption{Pyramidal RNN Embedding (PRE) Architecture.The actual pyramid structure is generated based on the settings.}
	\label{fig:PRE}
\end{figure}

In addition to sequentiality, time series data often exhibit multi-periodic characteristics. Convolutional Neural Networks, through multiple convolutional layers, can progressively construct larger-scale features, thereby capturing dependencies in time series data with long periods\cite{wu2022timesnet}. To capture multi-period dependencies in time series data, this paper proposes constructing a pyramidal multi-scale convolutional feature\cite{lin2017feature} for time series. The pyramidal structure consists of two parts: bottom-up and top-down. The bottom-up process constructs various cycles from small to large through one-dimensional convolution, representing the elevation of scale. In convolutional neural networks, higher layers of convolution often represent abstract semantic features, which in time series data imply changes in large-scale temporal cycles. Macroscopic changes are generally more deterministic than microscopic changes, making them easier to learn the underlying logic of changes\cite{hoel2013quantifying}. The top-down process gradually upsamples from the top layer downward. Since small-scale changes usually do not deviate too much from large-scale changes, this approach can transfer features from large cycles to small cycles as constraints. Finally, at each scale, the bottom-up and top-down features are added to obtain convolutional feature representations that are both finely detailed and constrained by macroscopic information.

Based on these considerations, this paper proposes an improved approach to Transformer prediction: first, construct pyramidal multi-scale convolutional features for univariate time series, then input the convolutional features of each scale into an RNN\cite{cho2014learning}  to learn multi-scale time series representations. By convoluting before using RNN, it can transform the original long sequence data into short sequence data, favorable for fully leveraging the advantages of RNN in modeling time series data. This method of learning univariate time series representations is named Pyramidal  Rnn Embedding(PRE). PRE is combined with the standard Transformer encoder, learning dependencies between multiple variables through a multi-head self-attention mechanism, and finally outputting predictions through linear projection, resulting in a model named PRformer.

Our contributions are as follows:

\begin{itemize}
	
	\item \textbf{Proposing a method for extracting multi-scale temporal representations.} We introduce a novel module called Pyramidal RNN Embedding (PRE), which learns multi-scale temporal features by combining a pyramidal structure with RNN. PRE captures temporal dependencies at different periods and scales, enabling it to obtain time series representations that fuse multi-scale dependency information. The computational complexity of PRE grows linearly with sequence length, ensuring high computational efficiency.

	\item \textbf{Proposing the use of PRE to enhance the performance of Transformer predictors.}
	We employ PRE to learn representations of univariate time series, which serve as input to the Transformer model. This approach effectively addresses the deficiencies of Transformers in capturing positional information relationships within sequences. Through extensive experiments combining PRE with three Transformer variants, we validate the significant improvement that PRE brings to Transformer-based predictors.
	
	\item \textbf{Proposing PRformer, a time series prediction model that combines PRE with the standard Transformer encoder.}
	By integrating PRE with the Transformer encoder, we introduce PRformer, a powerful time series prediction model. PRformer achieves state-of-the-art performance on several real-world datasets, surpassing existing Transformer-based architecture predictors and even outperforming linear prediction models. Moreover, we observe that PRformer's performance improves as the sequence length increases, highlighting its scalability and effectiveness in handling long-term dependencies.
	
\end{itemize}

\section{Related work}
\label{relatework}

\textbf{Transformer-Based Time Series Forecasting}
The Transformer has achieved widespread success in various tasks such as natural language processing \cite{vaswani2017attention} and computer vision \cite{dosovitskiy2021an}. In recent years, the Transformer has also been applied to time series forecasting with continuous improvements,such as Informer \cite{zhou2021informer},Autoformer \cite{wu2021autoformer} ,Pyraformer \cite{liu2021pyraformer},FedFormer \cite{zhou2022fedformer},PatchTST \cite{nie2023a} and so on. The improvements on the Transformer mainly involve designing new attention modules \cite{wu2021autoformer,zhou2022fedformer,liu2021pyraformer,zhou2021informer} and modifying the network architecture.

\textbf{RNN for Time Series Forecasting}
RNN has a strong affinity with sequence data in terms of its structure. With their gating mechanisms, RNN can adaptively regulate the influence of historical information by selectively depending on or ignoring past states.  For a significant period, RNN-based methods have held an important position in the field of time series forecasting \cite{lai2018modeling,salinas2020deepar,hewage2020temporal}. The use of RNN in time series forecasting helps capture the dependencies across different periods. GRU\cite{cho2014learning} controls the hidden state $h$ over time through reset and update gates, enabling it to learn "soft" periods. However, due to the fixed size of hidden units, RNN suffers from the gradient vanishing problem when modeling extremely long sequences. Nonetheless, RNN still possesses unique advantages, as its computational complexity is $O(nd^2)$,which is linear with respect to the length of the time series. Recently, some work try to  mitigate the gradient vanishing problem\cite{li2018independently} and applied RNN to long sequence modeling\cite{ma2023mega,peng2023rwkv}.

\textbf{Patch-Based Time Series Forecasting}
In computer vision, ViT\cite{dosovitskiy2021an} is a milestone work that segments images into 16x16 patches and feeds them into the Transformer model, resulting in significant success. Patch-based approaches have also been influential in natural language processing, speech, and other fields\cite{bao2021beit,hsu2021hubert}. In the domain of time series, PatchTST\cite{nie2023a} first divides time series into patches and feeds them into the Transformer model, achieving improved prediction accuracy compared to previous Transformer-based approaches. Patching is similar to one-dimensional convolution, allowing the extraction of local semantic information from the data. Moreover, patching combined with convolution can divide long sequences into shorter segments, which facilitates their modeling using RNN\cite{lai2018modeling,hewage2020temporal}.

\textbf{Multi-Scale Feature Extraction}
\cite{hoel2013quantifying} proposes that there are better causal relationships at a macroscopic level in terms of the causal changes in complex systems. This insight can be applied to time series forecasting, where incorporating features from the macroscopic level may lead to better predictions. In the field of image processing, feature pyramid networks\cite{lin2017feature} construct multi-scale features by combining convolution and upsampling, resulting in improved object detection accuracy. This concept also has important applications in weather forecasting, where constructing multi-scale features from storm data has been shown to enhance storm prediction\cite{yang2023customized}. In this paper, we propose constructing pyramidal convolutional features for univariate time series data to represent different time scales. RNN is applied to extract temporal representations from each scale, and the resulting representations are fused to obtain a multi-scale temporal representation, thereby improving time series forecasting performance.

\section{PRformer}
\label{3 method}
We primarily address the task of predicting multivariate time series, formalized as follows: Given a historical multivariate time series \(X \in \mathbb{R}^{L \times C}\) as input, the model aims to output future time series predictions \(Y \in \mathbb{R}^{H \times C}\). Here, \(C\) represents the number of variables in the time series, \(L\) signifies the time step length of the historical sequence, and \(H\) indicates the time step length for prediction.

\subsection{Model Architecture}
\label{3 Model Architecture}
\begin{figure*}
	\centering
	\includegraphics[width=0.8\textwidth]{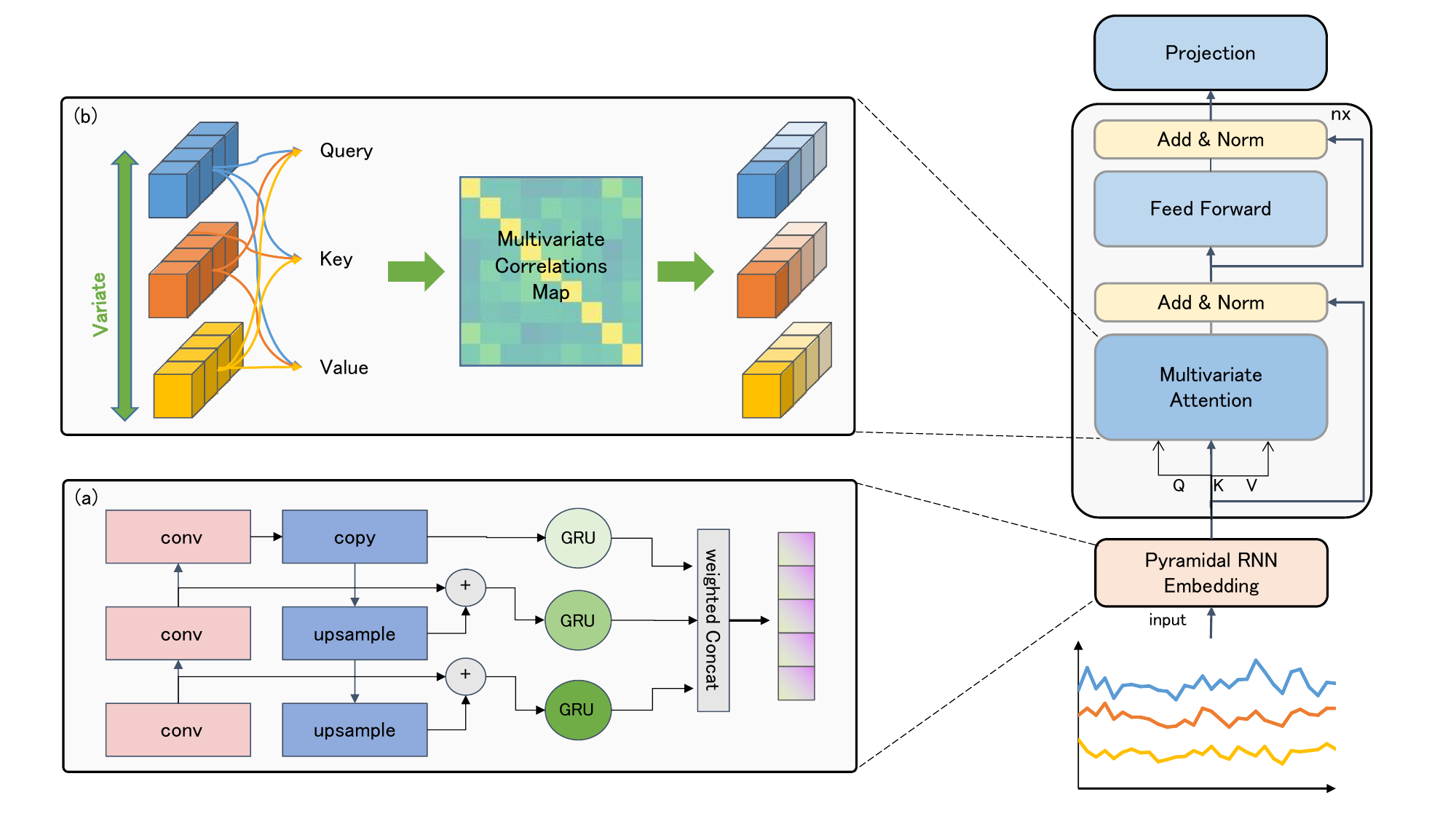}
	\caption{The overall architecture of PRformer utilizes a Transformer encoder as its backbone, culminating in the generation of prediction results through a simple linear projection. (a) Pyramidal RNN Embedding (PRE) block. The initial sequences of diverse variables are independently fed into PRE block to acquire distinct representations in the form of embeddings. (b) multi-head self-attention is employed on the embeddings of multiple variables to capture intricate interdependencies among them.}
	\label{fig:model_architecture}
\end{figure*}

The proposed PRformer is illustrated in Figure \ref{fig:model_architecture}.The model employs a vanilla Transformer encoder as its core architecture, comprising an embedding layer, multi-head attention, and a final projection layer. We introduce a Pyramid Recurrent Neural Network to extract embeddings for each univariate time series, emphasizing learning representations. The Transformer encoder focuses on learning relationships between multiple variables, obtaining representations for multivariate time series. Following the Transformer encoding, the final prediction results are obtained through a simple linear projection layer.

In the PRformer, the overall formalized process for predicting the future sequence \(Y \in \mathbb{R}^{H \times C}\) based on the historical sequence \(X \in \mathbb{R}^{L \times C}\) is expressed as follows:
\begin{equation}
	\begin{aligned}
		\mathbf{h}_{i}^{0} & = \mathrm{PREmbedding}(\mathbf{X}_{:,i}), \quad i=0,...,C-1 \\
		\mathbf{H}^{l+1} & = \mathrm{Encoder}(\mathbf{H}^{l}), \quad l=0,\cdots,L-1,  \\
		\hat{\mathbf{Y}}_{:,i} & = \mathrm{Projection}(\mathbf{h}_{i}^{L}), \quad i=0,...,C-1
	\end{aligned}
\end{equation}
Here, \(H = \{h_1, ..., h_{C}\} \in \mathbb{R}^{C \times D}\) represents embeddings for C variables, with each variable embedding having dimension D. The embeddings, generated through Pyramid RNN embedding(PRE), undergo processing in the Transformer encoder. In each encoder layer, the multi-head attention mechanism is employed to learn relationships between multiple variables and generate new representations. As PRE embedding already captures the sequential relationships of the sequence, additional positional embeddings are unnecessary for representing the entire sequence of a univariate variable.
\subsection{Pyramid RNN Embedding (PRE)}
The overall architecture of Pyramid RNN Embedding (PRE) is shown in Figure \ref{fig:PRE}. The input of PRE is a univariate time series, and the output is the embedded representation of the time series. The PRE consists of a Pyramid Temporal Convolution module and a Multi-Scale RNN module. The Pyramid Temporal Convolution module is used to learn multi-scale convolutional features of the temporal data, and compress the length of the time series through convolution, making it easier for the RNN to handle. The Multi-Scale RNN module is used to learn the sequence dependencies at different scales on the temporal dimension, and obtain the final multi-scale temporal embedding.

\subsubsection{Pyramid Convolution Block}
Periodicity is an important characteristic of time series data, and time series often have multiple different sized periods. In many cases, large periods contain small periods, such as weeks, months, seasons, and years. Previous time series predictors did not pay much attention to the relationship between large and small periods. In fact, there are dependencies between different-sized periods, and large periods may determine the fluctuation range of small periods, such as the influence of seasons on the climate. In addition, large-scale changes have less randomness compared to small-scale changes, making them less susceptible to noise interference and potentially yielding more robust prediction results. Based on this consideration, we propose the Pyramid Temporal Convolution module, and subsequent experiments have shown that it helps improve the performance of time series prediction.

\textbf{Bottom-up pathway} Through multiple layers of 1D convolution, a temporal feature pyramid is constructed, aligning each layer's data point period with the predetermined periodic lengths. Given the multiplicative growth pattern of the configured periodic lengths, the kernel size for each convolutional layer is equivalent to the multiple of the current layer's period and the subsequent layer's period length. The stride size for each layer's convolution is set equal to the kernel size. To facilitate downsampling, the output channels for all convolutional layers are maintained uniformly. The following equation illustrates the process of a univariate sequence \(x_{l-1}\) from the (l-1)-th layer through the one-dimensional convolutional kernel of the l-th layer.
\begin{equation}
	\begin{aligned}
		\mathbf{K}_l & = \left\lfloor \frac{\text{Window}^{l}}{\text{Window}^{l-1}} \right\rfloor \\
		\mathbf{x}_l & = \mathrm{Conv1d}(\mathbf{x}_{l-1}, \text{kernel}=\mathbf{K}_l, \text{stride}=\mathbf{K}_l)
	\end{aligned}
\end{equation}
\textbf{Top-down pathway and lateral connections} The top-down pathway samples from higher levels of the pyramid layers to obtain features with larger time scales and stronger semantics. Then, the same sized temporal features from the bottom-up pathway and the top-down pathway are fused through lateral connections. The features from the bottom-up pathway have smaller scale and higher resolution. Through fusion, the large-scale temporal changes can guide the small-scale changes, while maintaining a smaller time series scale to improve prediction accuracy. The following equation shows the process of generating the feature of the $(l-1)$th layer.
\begin{equation}
	\begin{aligned}
		\mathbf{x}_{l-1}^{'} & = \mathrm{upsample}(\mathbf{x}_{l}^{'}) \\
		\mathbf{x}_{l-1}^{\text{out}} & = \mathbf{x}_{l-1}^{\prime} + \mathbf{x}_{l-1}^{}
	\end{aligned}
\end{equation}
\subsubsection{Multi-Scale RNN Block}
The convolutional features from the Pyramid Convolution module still have a temporal structure for each scale, but the sequence length has been greatly reduced compared to the original sequence, which makes it easier to be processed by the RNN. The Multi-Scale RNN module is used to learn the sequence dependencies at different scales on the temporal dimension, and obtain the final multi-scale temporal embedding, where $D$ represents the embedding dimension of each variable. As shown in Figure \ref{fig:PRE}, the convolutional features of each scale are processed by a separate GRU module. The hidden dimension of the GRU is $D$ divided by the number of scales, and the last hidden unit is used as the temporal representation of that scale. The temporal representation $h^{(i)}$ of each scale is first multiplied by the weight coefficient $\beta_i$, and then concatenated to form the final multi-scale temporal embedding $h$. The weight coefficient $\beta_i$ is obtained by applying a softmax with temperature parameter control to $\alpha_i$, where $\alpha_i$ is initialized to $1/l$ and can be adaptively adjusted through training gradients. The temperature parameter $T$ of the softmax is used to amplify the differences in $\alpha_i$, making the coefficient $\beta_i$ sharper and increasing the differences in weight coefficients between different scale temporal representations.
\begin{equation}
	\begin{aligned}
		{h}^{(i)} & = \mathrm{GRU}({x}_{i}^{\text{out}}, D|\text{layer\_num}), \quad i=1,\ldots,l \\
		\beta_i & = \frac{\exp(\alpha_i/T)}{\sum_{j=1}^{l}\exp(\alpha_j/T)}, \quad i=1,\ldots,l \\
		{h} & = \mathrm{concat}(\beta_1 \cdot h^{(1)}, \beta_2 \cdot h^{(2)}, \ldots, \beta_l \cdot h^{(l)})
	\end{aligned}
\end{equation}

\subsubsection{Pyramid Convolutional Layer Configuration Method}

The configuration of pyramid convolutional layers is determined based on a set of periodic lengths, where each layer represents a distinct period. These layers extend from the bottom to the top, with periodic lengths increasing multiplicatively. The selection of periodic lengths is contingent upon the frequency of the time series dataset and the specific application scenario. For instance, in the case of the electrical power dataset ETTh1, with a temporal sampling frequency of 1 hour, one may opt for a set of periodic lengths such as 24, 48, and 96, corresponding to cycles of 1 day, 2 days, and 4 days, respectively. This configuration of the pyramid convolutional layers aligns more closely with real-world conditions, facilitating a more effective capture of cyclic dependencies.

\subsection{Transformer Encoder Multivariate Attention}
After the multivariate time series is processed by PRE, we obtain the embedded token $H = \{h_1,...,h_{C}\} \in \mathbb{R}^{C \times D}$, where $C$ represents the number of variables and $D$ represents the embedding dimension of each variable. The obtained multi-variate embedded token $H$ is then input into a regular Transformer encoder for processing. Each head in the multi-head attention mechanism linearly projects $H$ to obtain queries matrices $Q_{k}$, keys matrices $K_{k}$, and values matrices $V_{k}$, where $d_k$ represents the projected dimension of each head. Then, a scaled dot product is applied to obtain the attention output $\mathbf{O}_k \in \mathbb{R}^{C \times d_k}$:

\begin{equation}
	(\mathbf{O}_{k})^{T} = \mathrm{Softmax}\left(\frac{Q_{k}K_{k}^{T}}{\sqrt{d_{k}}}\right)V_{k}
\end{equation}

The multi-head attention block also includes LayerNorm layers and a feed forward network with residual connections. It generates the representation $z^{(i)} \in \mathbb{R}^{C \times D}$ for each variable. In each Encoder layer, the multi-head attention mechanism is able to learn the relationships between different variables and generate new representations for each variable. Finally, a channel-wise linear projection is applied to obtain the prediction results $(\hat{y}_{L+1}^{(i)},...,\hat{y}_{L+T}^{(i)}) \in \mathbb{R}^{1 \times H}$ for each variable. The prediction results are concatenated to obtain the final prediction result $\hat{Y} \in \mathbb{R}^{H \times C}$.

\subsection{Loss Function and Normalization}
We choose to use the MAE loss function to train the model. The MAE loss function is defined as follows:

\begin{equation}
	\mathcal{L}(Y,\bar{Y}) = \frac{1}{HC}\sum_{t=1}^{H}\sum_{i=1}^{C}|\bar{y}_{t}^{(i)} - y_{t}^{(i)}|
\end{equation}

\textbf{Instance Normalization.} In this paper, we use reversible instance normalization\cite{kim2022reversible}(RevIN)to process the time series data to alleviate distribution shift issues. Similar to standardization, RevIN scales the sequence data of each variable to $N(\beta_i,\gamma_i)$ through learnable parameters. RevIN is applied to normalize the time series data before entering the model, and then applied again to reverse normalize the predicted results to obtain the final prediction results.

\subsection{Complexity Analysis of PRformer}

\begin{table}[htbp]
	\centering
	\caption{Complexity analysis of different Transformer prediction models. S is the stride length of PatchTST.}
	\label{tab:complexity}
	\resizebox{0.48\textwidth}{!}{
		\begin{tabular}{lcc}
			\hline
			\textbf{Methods} & \textbf{Time} & \textbf{Memory} \\
			\hline
			Transformer & $O(L^2)$ & $O(L^2)$ \\
			Informer & $O(L\log L)$ & $O(L\log L)$ \\
			Reformer & $O(L\log L)$ & $O(L\log L)$ \\
			PatchTST & $O(\frac{L^2}{S^2})$ & $O(\frac{L^2}{S^2})$ \\
			PRformer & $O(\frac{L}{W}+D^2)$ & $O(\frac{L}{W}+D^2)$ \\
			PRE+Informer & $O(\frac{L}{W}+D\log D)$ & $O(\frac{L}{W}+D\log D)$ \\
			PRE+Reformer & $O(\frac{L}{W}+D\log D)$ & $O(\frac{L}{W}+D\log D)$ \\
			\hline
		\end{tabular}
	}
\end{table}

The PRformer primarily consists of two components: the PRE and the Transformer encoder. The time complexity of the PRE is related to the length of the historical sequence and is $O(\frac{L}{W})$, where W is the period length of the bottom layer in the pyramid structure. For some real-world data recorded in hours, W is typically set to 24, representing a daily period, significantly reducing the time complexity. The space complexity of PRE is associated with the dimensionality of the embedding, D, and is $O(D)$. Both the time and space complexities of the Transformer encoder are $O(D^2)$. Typically, D is chosen such that $ D \leq L$, hence the overall time complexity of the PRformer is $O(\frac{L}{W} + D^2)$ and the space complexity is $O(\frac{L}{W} + D^2)$. The table \ref{tab:complexity} compares the time complexity and memory usage of various Transformer models. Compared to the state-of-the-art Transformer predictor model, PatchTST, and the original Transformer, the PRformer has lower complexity. It is noteworthy that the complexity of the PRformer grows linearly with sequence length, enabling it to utilize longer lookback windows under equivalent hardware conditions, thereby enhancing predictive performance. Furthermore, the key module of PRformer, PRE, can be combined with Informer or Reformer to further reduce complexity and memory usage. Experiments in section \ref{sec:exp} show that such combinations can also significantly improve the predictive performance of the original models.

\section{EXPERIMENTS}
\label{sec:exp}%

In various time series forecasting applications, we comprehensively evaluate PRformer to validate the generalization of the proposed model. We also assess the effectiveness of the pyramidal RNN embedding layer in enhancing various Transformer predictors, confirming the ability of the pyramidal RNN embedding to learn effective representations of univariate time series.

\textbf{Datasets} We assess the performance of PRformer on 8 popular datasets, including Weather, Traffic, Electricity, Solar-Energy and 4 ETT datasets (ETTh1, ETTh2, ETTm1, ETTm2). These datasets represent diverse domains and intervals, offering a comprehensive assessment environment.The characteristics of these datasets are summarized in Table \ref{tab:dataset-summary}. Of particular note are the Electricity and Traffic datasets, which consist of over 300 variables. For such datasets, modeling the dependencies between multiple variables is crucial, and our proposed PRformer excels in this aspect.

\begin{table}[htbp]
	\centering
	\caption{Summary of datasets.}
	\resizebox{0.48\textwidth}{!}{
		\begin{tabular}{ccccccccc}
			\toprule
			Datasets & Weather & Traffic & Electricity & ETTh1 & ETTh2 & ETTm1 & ETTm2 & Solar-Energy \\
			\midrule
			Channels & 21 & 862 & 321 & 7 & 7 & 7 & 7 & 137 \\
			Frequency & 10 mins & 1 hour & 1 hour & 1 hour & 1 hour & 15 mins & 15 mins & 10 mins\\
			Timesteps & 52,696 & 17,544 & 26,304 & 17,420 & 17,420 & 69,680 & 69,680 & 36601\\
			\bottomrule
		\end{tabular}%
	}
	\label{tab:dataset-summary}%
\end{table}%

\textbf{Baselines and metrics}
We select SOTA Transformer-based models and some representative non-Transformer-based model in the time series forecasting field as baselines, including iTransformer \cite{liu2024itransformer}, PatchTST \cite{nie2023a}, Crossformer \cite{zhang2022crossformer}, FEDformer \cite{zhou2022fedformer}, Autoformer \cite{wu2021autoformer}, Informer \cite{zhou2021informer},Dlinear \cite{zeng2023are} and TimesNet \cite{wu2022timesnet}.Two commonly used evaluation metrics, Mean Squared Error (MSE) and Mean Absolute Error (MAE), serve as the assessment criteria.

\begin{table*}[htbp]
	\centering
	\caption{Performance promotion obtained by PRE.}
	\resizebox{0.8\textwidth}{!}{
		\begin{tabular}{c|c|cc|cc|cc|cc}
			\toprule
			\multicolumn{2}{c|}{Models} & \multicolumn{2}{p{10.165em}|}{Transformer\newline{}(2017)} & \multicolumn{2}{p{8.22em}|}{Reformer\newline{}(2020)} & \multicolumn{2}{p{8.22em}|}{Informer\newline{}(2021)} & \multicolumn{2}{p{8.22em}}{Flowformer\newline{}(2022)} \\
			\cmidrule{3-10}    \multicolumn{2}{c|}{Metric} & MSE   & MAE   & MSE   & MAE   & MSE   & MAE   & MSE   & MAE \\
			\midrule
			\multirow{4}[4]{*}{Electricity} & Original & 0.277  & 0.372  & 0.338  & 0.422  & 0.311  & 0.397  & 0.267  & 0.359  \\
			& \textbf{+PRE} & 0.156  & 0.247  & 0.163  & 0.258  & 0.162  & 0.257  & 0.161  & 0.256  \\
			\cmidrule{2-10}          & Promotion & 43.86\% & 33.74\% & 51.80\% & 38.98\% & 47.92\% & 35.38\% & 39.63\% & 28.81\% \\
			\midrule
			\multirow{4}[4]{*}{Traffic} & Original & 0.665  & 0.363  & 0.741  & 0.422  & 0.764  & 0.416  & 0.750  & 0.421  \\
			& \textbf{+PRE} & 0.383  & 0.239  & 0.394  & 0.273  & 0.394  & 0.273  & 0.392  & 0.272  \\
			\cmidrule{2-10}          & Promotion & 42.44\% & 34.30\% & 46.81\% & 35.23\% & 48.46\% & 34.44\% & 47.79\% & 35.28\% \\
			\midrule
			\multirow{4}[4]{*}{Weather} & Original & 0.657 & 0.572 & 0.803 & 0.656 & 0.634 & 0.548 & 0.286 & 0.308 \\
			& \textbf{+PRE} & 0.225 & 0.257 & 0.224 & 0.264 & 0.231 & 0.269 & 0.227 & 0.268 \\
			\cmidrule{2-10}          & Promotion & 65.79\% & 55.16\% & 72.14\% & 59.79\% & 63.61\% & 50.84\% & 20.63\% & 12.92\% \\
			\toprule
			\multicolumn{2}{c|}{Promotion Average} & \multicolumn{2}{c|}{45.88\%} & \multicolumn{2}{c|}{50.79\%} & \multicolumn{2}{c|}{46.78\%} & \multicolumn{2}{c}{30.84\%} \\
			\bottomrule
			
		\end{tabular}%
		\label{tab:promotion_avg}%
	}
\end{table*}%

\subsection{Performance promotion with PRE}
To validate the effectiveness and generalization capability of PRE embedding, we apply PRE to Transformer\cite{vaswani2017attention} and its variants (primarily addressing the quadratic complexity issue of the self-attention mechanism): Reformer \cite{kitaev2020reformer}, Informer\cite{zhou2021informer},and  Flowformer \cite{wu2022flowformer}. The experimental results reveal surprising and promising findings.Table \ref{tab:promotion_avg} present the performance improvements achieved by applying PRE. Notably, PRE consistently enhances the performance of various Transformer variants. Overall, a performance improvement of 45.88\% is achieved on the Transformer, 50.79\% on the Reformer, 46.78\% on the Informer, and 30.84\% on the Flowformer. This indicates that PRE can effectively represent univariate time series learning and enhance the performance of Transformer-based predictors. In the future, the concept of PRE can be widely practiced in Transformer-based predictors to take advantage of the growing efficient attention mechanisms.

\subsection{Long-term Time Series Forecasting}

The forecasting results for multivariate long-term time series are presented in Table \ref{tab:lstfResults}, with the best outcomes highlighted in bold and the second-best underlined. The proposed PRformer model has achieved state-of-the-art performance across most scenarios, securing the top performance in 25 out of 32 MSE metrics and in 28 out of 32 MAE metrics. Even when compared to the previously state-of-the-art Transformer-based forecaster, PatchTST, the PRformer demonstrated comprehensive performance improvements. This enhancement in performance was particularly notable in high-dimensional time series datasets, such as the Traffic and Solar-Energy datasets. Against the Crossformer model, which explicitly captures multivariate correlations, the PRformer's performance improvement exceeded 70\% in some cases, showcasing its effectiveness in modeling dependencies among multiple variables. We attribute this efficiency to the use of PRE  for time series representation learning, which alleviates the limitations of Transformer positional encodings, thereby leading to improved performance. This is also why the PRformer exhibited comprehensive advantages over advanced linear models and CNN-based models, underscoring the success of the PRformer design.

\begin{table*}[htbp]
	\centering
	\caption{Results for multivariate long-term series forecasting on seven datasets with input length \(L = 720\) and prediction length \(H \in (96,192,336,720)\). A lower value indicates better performance, the best results are highlighted in bold and the second are underlined.}
	\resizebox{1\textwidth}{!}{
		\begin{tabular}{c|c|cc|cc|cc|cc|cc|cc|cc|cc|cc}
			\toprule
			\multicolumn{2}{p{7.13em}|}{Models} & \multicolumn{2}{p{7.13em}|}{PRformer\newline{}(2024)} & \multicolumn{2}{p{7.13em}|}{iTransformer\newline{}(2024)} & \multicolumn{2}{p{7.13em}|}{PatchTST\newline{}(2023)} & \multicolumn{2}{p{7.13em}|}{Crossformer\newline{}(2023)} & \multicolumn{2}{p{7.13em}|}{FEDformer\newline{}(2022)} & \multicolumn{2}{p{7.13em}|}{Autoformer\newline{}(2021)} & \multicolumn{2}{p{7.13em}|}{Informer\newline{}(2021)} & \multicolumn{2}{p{7.13em}|}{Dlinear\newline{}(2023)} & \multicolumn{2}{p{7.13em}}{TimesNet\newline{}(2023)} \\
			\midrule
			\multicolumn{2}{c|}{Metric} & MSE   & MAE   & MSE   & MAE   & MSE   & MAE   & MSE   & MAE   & MSE   & MAE   & MSE   & MAE   & MSE   & MAE   & MSE   & MAE \\
			\midrule
			\multirow{4}{*}{ETTm1} & 96    & \textbf{0.278 } & \textbf{0.333 } & 0.334  & 0.368 & \underline{0.293}  & 0.346  & 0.404  & 0.426  & 0.326  & 0.390  & 0.505  & 0.475  & 0.626  & 0.560  & 0.299  & \underline{0.343}  & 0.338  & 0.375  \\
			& 192   & \textbf{0.324 } & \textbf{0.361 } & 0.377  & 0.391 & \underline{0.333}  & 0.370  & 0.450  & 0.451  & 0.365  & 0.415  & 0.553  & 0.496  & 0.725  & 0.619  & 0.335  & \underline{0.365}  & 0.374  & 0.387  \\
			& 336   & \textbf{0.362 } & \textbf{0.384 } & 0.426  & 0.420 & \underline{0.369}  & 0.392  & 0.532  & 0.515  & 0.392  & 0.425  & 0.621  & 0.537  & 1.005  & 0.741  & \underline{0.369}  & \underline{0.386}  & 0.410  & 0.411  \\
			& 720   & 0.426  & 0.425  & 0.491  & 0.459 & \textbf{0.416 } & \textbf{0.420 } & 0.666  & 0.589  & 0.446  & 0.458  & 0.671  & 0.561  & 1.133  & 0.845  & \underline{0.425}  & \underline{0.421}  & 0.478  & 0.450  \\
			\midrule
			\multirow{4}{*}{ETTm2} & 96    & \textbf{0.162 } & \textbf{0.245 } & 0.180  & 0.264 & \underline{0.166}  & \underline{0.256}  & 0.287  & 0.366  & 0.180  & 0.271  & 0.255  & 0.339  & 0.355  & 0.462  & 0.167  & 0.260  & 0.187  & 0.267  \\
			& 192   & \textbf{0.219 } & \textbf{0.286 } & 0.250  & 0.309 & \underline{0.223}  & \underline{0.296}  & 0.414  & 0.492  & 0.252  & 0.318  & 0.281  & 0.340  & 0.595  & 0.586  & 0.224  & 0.303  & 0.249  & 0.309  \\
			& 336   & \textbf{0.272 } & \textbf{0.326 } & 0.311  & 0.348 & \underline{0.274}  & \underline{0.329}  & 0.597  & 0.542  & 0.324  & 0.364  & 0.339  & 0.372  & 1.270  & 0.871  & 0.281  & 0.342  & 0.321  & 0.351  \\
			& 720   & \textbf{0.359 } & \textbf{0.383 } & 0.412  & 0.407 & \underline{0.362}  & \underline{0.385}  & 1.730  & 1.042  & 0.410  & 0.420  & 0.422  & 0.419  & 3.001  & 1.267  & 0.397  & 0.421  & 0.408  & 0.403  \\
			\midrule
			\multirow{4}{*}{ETTh1} & 96    & \textbf{0.354 } & \textbf{0.383 } & 0.386  & 0.405 & \underline{0.370} & 0.400  & 0.423  & 0.448  & 0.376  & 0.415  & 0.449  & 0.459  & 0.941  & 0.769  & 0.375  & \underline{0.399}  & 0.384  & 0.402  \\
			& 192   & \textbf{0.397 } & \textbf{0.410 } & 0.441  & 0.436 & 0.413  & 0.429  & 0.471  & 0.474  & 0.423  & 0.446  & 0.500  & 0.482  & 1.007  & 0.786  & \underline{0.405}  & \underline{0.416}  & 0.436  & 0.429  \\
			& 336   & \underline{0.427}  & \textbf{0.428 } & 0.487  & 0.458 & \textbf{0.422 } & \underline{0.440}  & 0.570  & 0.546  & 0.444  & 0.462  & 0.521  & 0.496  & 1.038  & 0.784  & 0.439  & 0.443  & 0.491  & 0.469  \\
			& 720   & 0.489  & 0.492 & 0.503  & 0.491 & \textbf{0.447 } & \textbf{0.468 } & 0.653  & 0.621  & \underline{0.469}  & 0.492  & 0.514  & 0.512  & 1.144  & 0.857  & 0.472  & \underline{0.490}  & 0.521  & 0.500  \\
			\midrule
			\multirow{4}[2]{*}{ETTh2} & 96    & \textbf{0.268 } & \textbf{0.327 } & 0.297  & 0.349  & \underline{0.274}  & \underline{0.337}  & 0.745  & 0.584  & 0.332  & 0.374  & 0.358  & 0.397  & 1.549  & 0.952  & 0.289  & 0.353  & 0.340  & 0.374  \\
			& 192   & \textbf{0.332 } & \textbf{0.370 } & 0.380  & 0.400 & \underline{0.341}  & \underline{0.382}  & 0.877  & 0.656  & 0.407  & 0.446  & 0.456  & 0.452  & 3.792  & 1.542  & 0.383  & 0.418  & 0.402  & 0.414  \\
			& 336   & \underline{0.361}  & \underline{0.395} & 0.428  & 0.432  & \textbf{0.329 } & \textbf{0.384 } & 1.043  & 0.731  & 0.400  & 0.447  & 0.482  & 0.486  & 4.215  & 1.642  & 0.448  & 0.465  & 0.452  & 0.452  \\
			& 720   & \underline{0.396}  & \underline{0.429} & 0.427  & 0.445 & \textbf{0.379 } & \textbf{0.422 } & 1.104  & 0.763  & 0.412  & 0.469  & 0.515  & 0.511  & 3.656  & 1.619  & 0.605  & 0.551  & 0.462  & 0.468  \\
			\midrule
			\multirow{4}[2]{*}{Electricity} & 96    & \textbf{0.127 } & \textbf{0.217 } & 0.148  & 0.240 & \underline{0.129}  & \underline{0.222}  & 0.151  & 0.251  & 0.186  & 0.302  & 0.201  & 0.317  & 0.304  & 0.393  & 0.140  & 0.237  & 0.168  & 0.272  \\
			& 192   & \underline{0.148}  & \textbf{0.237 } & 0.162  & 0.253 & \textbf{0.147 } & \underline{0.240}  & 0.163  & 0.262  & 0.197  & 0.311  & 0.222  & 0.334  & 0.327  & 0.417  & 0.153  & 0.249  & 0.184  & 0.289  \\
			& 336   & \textbf{0.161 } & \textbf{0.252 } & 0.178  & 0.269 & \underline{0.163}  & \underline{0.259}  & 0.195  & 0.288  & 0.213  & 0.328  & 0.231  & 0.338  & 0.333  & 0.422  & 0.169  & 0.267  & 0.198  & 0.300  \\
			& 720   & \textbf{0.185 } & \textbf{0.275 } & 0.225  & 0.317 & \underline{0.197}  & \underline{0.290}  & 0.224  & 0.316  & 0.233  & 0.344  & 0.254  & 0.361  & 0.351  & 0.427  & 0.203  & 0.301  & 0.220  & 0.320  \\
			\midrule
			\multirow{4}[2]{*}{Traffic} & 96    & \textbf{0.353 } & \textbf{0.222 } & 0.395  & 0.268 & \underline{0.360}  & \underline{0.249}  & 0.522  & 0.290  & 0.576  & 0.359  & 0.613  & 0.388  & 0.733  & 0.410  & 0.410  & 0.282  & 0.593  & 0.321  \\
			& 192   & \textbf{0.372 } & \textbf{0.233 } & 0.417  & 0.276 & \underline{0.379}  & \underline{0.256}  & 0.530  & 0.293  & 0.610  & 0.380  & 0.616  & 0.382  & 0.777  & 0.435  & 0.423  & 0.287  & 0.617  & 0.336  \\
			& 336   & \textbf{0.385 } & \textbf{0.241 } & 0.433  & 0.283 & \underline{0.392}  & \underline{0.264}  & 0.558  & 0.305  & 0.608  & 0.375  & 0.622  & 0.337  & 0.776  & 0.434  & 0.436  & 0.296  & 0.629  & 0.336  \\
			& 720   & \textbf{0.421 } & \textbf{0.258 } & 0.467  & 0.302 & \underline{0.432}  & \underline{0.286}  & 0.589  & 0.328  & 0.621  & 0.375  & 0.660  & 0.408  & 0.827  & 0.466  & 0.466  & 0.315  & 0.640  & 0.350  \\
			\midrule
			\multirow{4}[2]{*}{Weather} & 96    & \textbf{0.144 } & \textbf{0.187 } & 0.174  & 0.214 & \underline{0.149}  & \underline{0.198}  & 0.158  & 0.230  & 0.238  & 0.314  & 0.266  & 0.336  & 0.354  & 0.405  & 0.176  & 0.237  & 0.172  & 0.220  \\
			& 192   & \textbf{0.188 } & \textbf{0.233 } & 0.221  & 0.254 & \underline{0.194}  & \underline{0.241}  & 0.206  & 0.277  & 0.275  & 0.329  & 0.307  & 0.367  & 0.419  & 0.434  & 0.220  & 0.282  & 0.219  & 0.261  \\
			& 336   & \textbf{0.241 } & \textbf{0.274 } & 0.278  & 0.296 & \underline{0.245}  & \underline{0.282}  & 0.272  & 0.335  & 0.339  & 0.377  & 0.359  & 0.395  & 0.583  & 0.543  & 0.265  & 0.319  & 0.280  & 0.306  \\
			& 720   & 0.326  & \textbf{0.332 } & 0.358  & 0.349 & \textbf{0.314 } & \underline{0.334}  & 0.398  & 0.418  & 0.389  & 0.409  & 0.419  & 0.428  & 0.916  & 0.705  & \underline{0.323}  & 0.362  & 0.365  & 0.359  \\
			\midrule
			\multirow{4}[2]{*}{Solar-Energy} & 96    & \textbf{0.171 } & \textbf{0.201 } & \underline{0.203}  & 0.237 & 0.234  & 0.286  & 0.310  & 0.331  & 0.242  & 0.342  & 0.884  & 0.711  & \underline{0.206}  & \underline{0.229}  & 0.290  & 0.378  & 0.250  & 0.292  \\
			& 192   & \textbf{0.193 } & \textbf{0.215 } & \underline{0.233}  & 0.261  & 0.267  & 0.310  & 0.734  & 0.725  & 0.285  & 0.380  & 0.834  & 0.692  & \underline{0.235}  & \underline{0.258}  & 0.320  & 0.398  & 0.296  & 0.318  \\
			& 336   & \textbf{0.217 } & \textbf{0.228 } & \underline{0.248}  & 0.273 & 0.290  & 0.315  & 0.750  & 0.735  & 0.282  & 0.376  & 0.941  & 0.723  & \underline{0.261}  & \underline{0.271}  & 0.353  & 0.415  & 0.319  & 0.330  \\
			& 720   & \textbf{0.229 } & \textbf{0.237 } & \underline{0.249}  & 0.275 & 0.289  & 0.317  & 0.769  & 0.765  & 0.357  & 0.427  & 0.882  & 0.717  & \underline{0.264}  & \underline{0.279}  & 0.356  & 0.413  & 0.338  & 0.337  \\
			\midrule
			\multicolumn{2}{c|}{$1^{st}$ Count} & 25   & 28   & 0   & 0  & 7   & 4   & 0   & 0   & 0   & 0   & 0   & 0   & 0   & 0   & 0   & 0   & 0   & 0 \\
			
			\bottomrule

		\end{tabular}%
	}
	\label{tab:lstfResults}%
\end{table*}%

\subsection{Model Analysis}

\textbf{Ablation Experiment}
To validate the effectiveness of the components of PRformer, we conducted ablation experiments. Three variants of PRformer were tested: (1) PRformer V1: we replaced the multi-head self-attention module in the Transformer with a linear layer, (2) PRformer V2: we replaced the generation of time series representations for each variable with a linear projection instead of PRE, (3) PRformer V3: we only used the bottom layer of the temporal convolution pyramid to extract univariate time series representations. We used PatchTST as the state-of-the-art benchmark for the Transformer-based model. If the ablated version outperformed PatchTST, it is indicated in bold numbers. From Table \ref{tab:ablation}, it can be observed that 12/32 instances of PRformer V2 showed improvement, 18/32 instances of PRformer V1 showed improvement, and 23/32 instances of PRformer V3 showed improvement. The best performance was achieved by the PRformer model proposed in this paper, with 26/32 instances showing improvement. By comparing PRformer V2 and PRformer, we validated the effectiveness of the PRE embedding compared to a simple linear layer; by comparing PRformer V1 and PRformer, we confirmed the effectiveness of the Transformer's self-attention module for learning relationships between multiple variables; by comparing PRformer V3 and PRformer, we verified the effectiveness of building multiscale temporal representations through the pyramid structure.

\begin{table*}[htbp]
	\centering
	\caption{Ablation studies: multivariate long-term series forecasting result with input length \(L=720\) and prediction length \(H \in (96,192,336,720)\).Evaluation and comparison of three PRformer variants against baseline across datasets: electricity, traffic, weather, and ETTh2. Instances of enhanced performance over baseline are emphasized in bold.}
	\resizebox{0.7\textwidth}{!}{
		\begin{tabular}{cc|cc|cc|cc|cc|cc}
			\toprule
			\multicolumn{2}{c|}{\multirow{2}[4]{*}{Models}} & \multicolumn{8}{c|}{PRformer}                                 & \multicolumn{2}{c}{\multirow{2}[4]{*}{PatchTST\newline{}(2023)}} \\
			\cmidrule{3-10}    \multicolumn{2}{c|}{} & \multicolumn{2}{c|}{PRformer} & \multicolumn{2}{c|}{PRformer V1} & \multicolumn{2}{c|}{PRformer V2} & \multicolumn{2}{c|}{PRformer V3} & \multicolumn{2}{c}{} \\
			\midrule
			\multicolumn{2}{c|}{Metric} & MSE   & MAE   & MSE   & MAE   & MSE   & MAE   & MSE   & MAE   & MSE   & MAE \\
			\midrule
			\multicolumn{1}{c|}{\multirow{4}[2]{*}{Electricity}} & 96    & \textbf{0.127 } & \textbf{0.217 } & 0.132  & \textbf{0.219 } & 0.132  & 0.222  & 0.130  & \textbf{0.218 } & 0.129  & 0.222  \\
			\multicolumn{1}{c|}{} & 192   & 0.148  & \textbf{0.237 } & 0.147  & \textbf{0.235 } & 0.149  & 0.240  & 0.148  & \textbf{0.237 } & 0.147  & 0.240  \\
			\multicolumn{1}{c|}{} & 336   & \textbf{0.161 } & \textbf{0.252 } & 0.165  & \textbf{0.254 } & \textbf{0.161 } & \textbf{0.252 } & \textbf{0.161 } & \textbf{0.253 } & 0.163  & 0.259  \\
			\multicolumn{1}{c|}{} & 720   & \textbf{0.185 } & \textbf{0.275 } & \textbf{0.196 } & \textbf{0.283 } & \textbf{0.176 } & \textbf{0.269 } & \textbf{0.190 } & \textbf{0.279 } & 0.197  & 0.290  \\
			\midrule
			\multicolumn{1}{c|}{\multirow{4}[2]{*}{Traffic}} & 96    & \textbf{0.353 } & \textbf{0.222 } & \textbf{0.358 } & \textbf{0.230 } & \textbf{0.344 } & \textbf{0.227 } & \textbf{0.344 } & \textbf{0.222 } & 0.360  & 0.249  \\
			\multicolumn{1}{c|}{} & 192   & \textbf{0.372 } & \textbf{0.233 } & \textbf{0.377 } & \textbf{0.238 } & \textbf{0.370 } & \textbf{0.239 } & \textbf{0.372 } & \textbf{0.233 } & 0.379  & 0.256  \\
			\multicolumn{1}{c|}{} & 336   & \textbf{0.385 } & \textbf{0.241 } & 0.397  & \textbf{0.248 } & \textbf{0.385 } & \textbf{0.247 } & \textbf{0.382 } & \textbf{0.239 } & 0.392  & 0.264  \\
			\multicolumn{1}{c|}{} & 720   & \textbf{0.421 } & \textbf{0.258 } & 0.442  & \textbf{0.271 } & \textbf{0.425 } & \textbf{0.265 } & \textbf{0.412 } & \textbf{0.257 } & 0.432  & 0.286  \\
			\midrule
			\multicolumn{1}{c|}{\multirow{4}[2]{*}{Weather}} & 96    & \textbf{0.144 } & \textbf{0.187 } & \textbf{0.145 } & \textbf{0.183 } & 0.162  & 0.202  & 0.152  & \textbf{0.191 } & 0.149  & 0.198  \\
			\multicolumn{1}{c|}{} & 192   & \textbf{0.188 } & \textbf{0.233 } & \textbf{0.193 } & \textbf{0.230 } & 0.205  & 0.248  & 0.196  & \textbf{0.236 } & 0.194  & 0.241  \\
			\multicolumn{1}{c|}{} & 336   & \textbf{0.241 } & \textbf{0.274 } & 0.250  & \textbf{0.275 } & 0.251  & 0.284  & \textbf{0.243 } & \textbf{0.275 } & 0.245  & 0.282  \\
			\multicolumn{1}{c|}{} & 720   & 0.326  & \textbf{0.332 } & 0.355  & 0.343  & 0.357  & 0.349  & 0.318  & \textbf{0.329 } & 0.314  & 0.334  \\
			\midrule
			\multicolumn{1}{c|}{\multirow{4}[2]{*}{ETTh2}} & 96    & \textbf{0.268 } & \textbf{0.327 } & 0.278  & \textbf{0.328 } & 0.295  & 0.353  & \textbf{0.267 } & \textbf{0.327 } & 0.274  & 0.337  \\
			\multicolumn{1}{c|}{} & 192   & \textbf{0.332 } & \textbf{0.370 } & 0.341  & \textbf{0.371 } & 0.360  & 0.393  & \textbf{0.338 } & \textbf{0.373 } & 0.341  & 0.382  \\
			\multicolumn{1}{c|}{} & 336   & 0.361  & 0.395  & 0.366  & 0.396  & 0.396  & 0.419  & 0.368  & 0.399  & 0.329  & 0.384  \\
			\multicolumn{1}{c|}{} & 720   & 0.396  & 0.429  & 0.408  & 0.434  & 0.446  & 0.467  & 0.397  & 0.428  & 0.379  & 0.422  \\
			\midrule
			\multicolumn{2}{c|}{count} & \multicolumn{2}{c|}{26} & \multicolumn{2}{c|}{18} & \multicolumn{2}{c|}{12} & \multicolumn{2}{c|}{23} & \multicolumn{2}{c}{-} \\
			\bottomrule
		\end{tabular}%
	}
	\label{tab:ablation}%
\end{table*}%

\textbf{Lookback Window Analysis}

\begin{figure}[!t]
	\centering
	\vspace{0.1cm} 
	\begin{minipage}{1\columnwidth}
		\centering
		\subfloat[]{\includegraphics[width=0.48\linewidth]{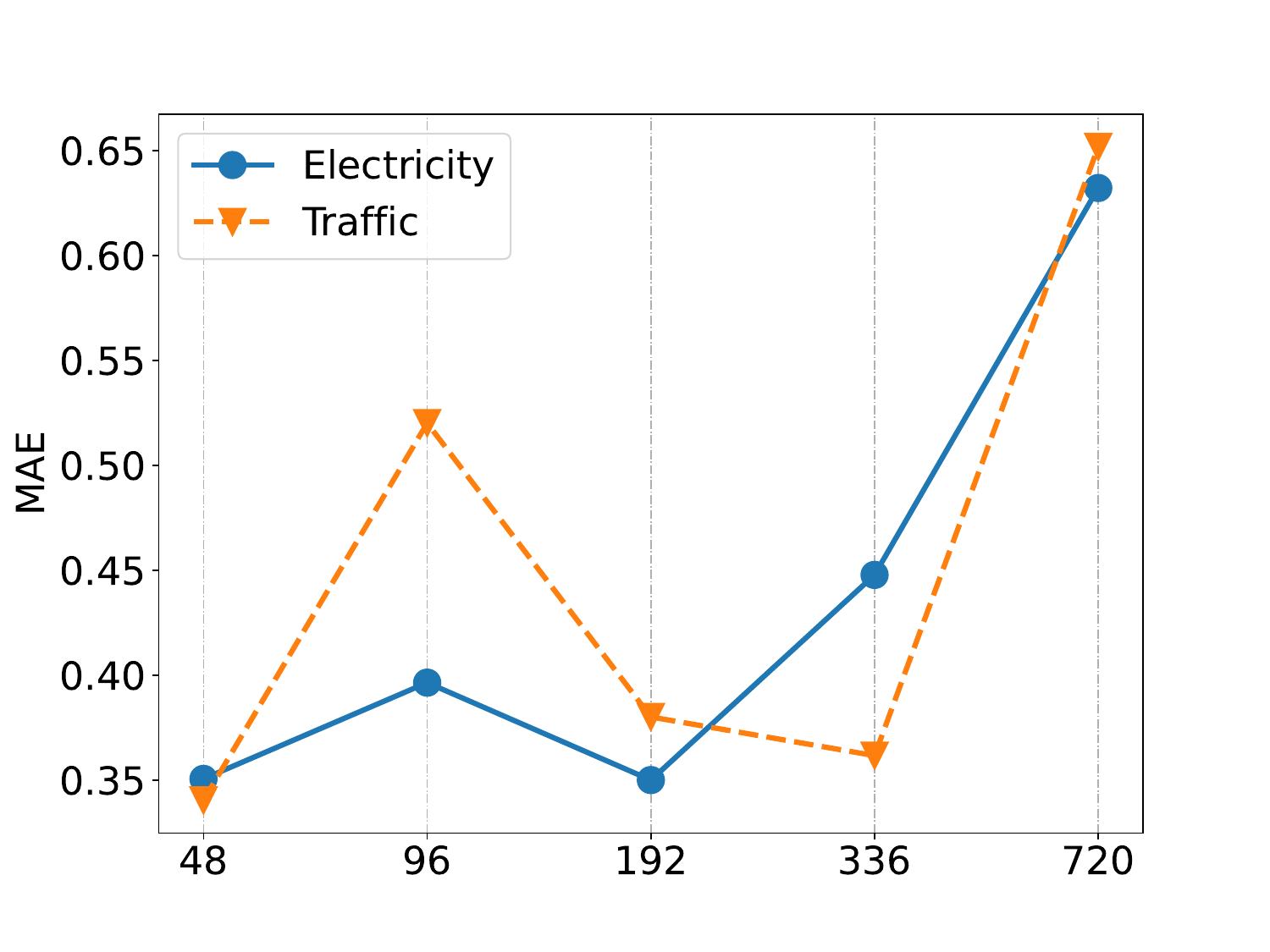}%
			\label{fig:lookback.1}}
		\hfill
		\subfloat[]{\includegraphics[width=0.48\linewidth]{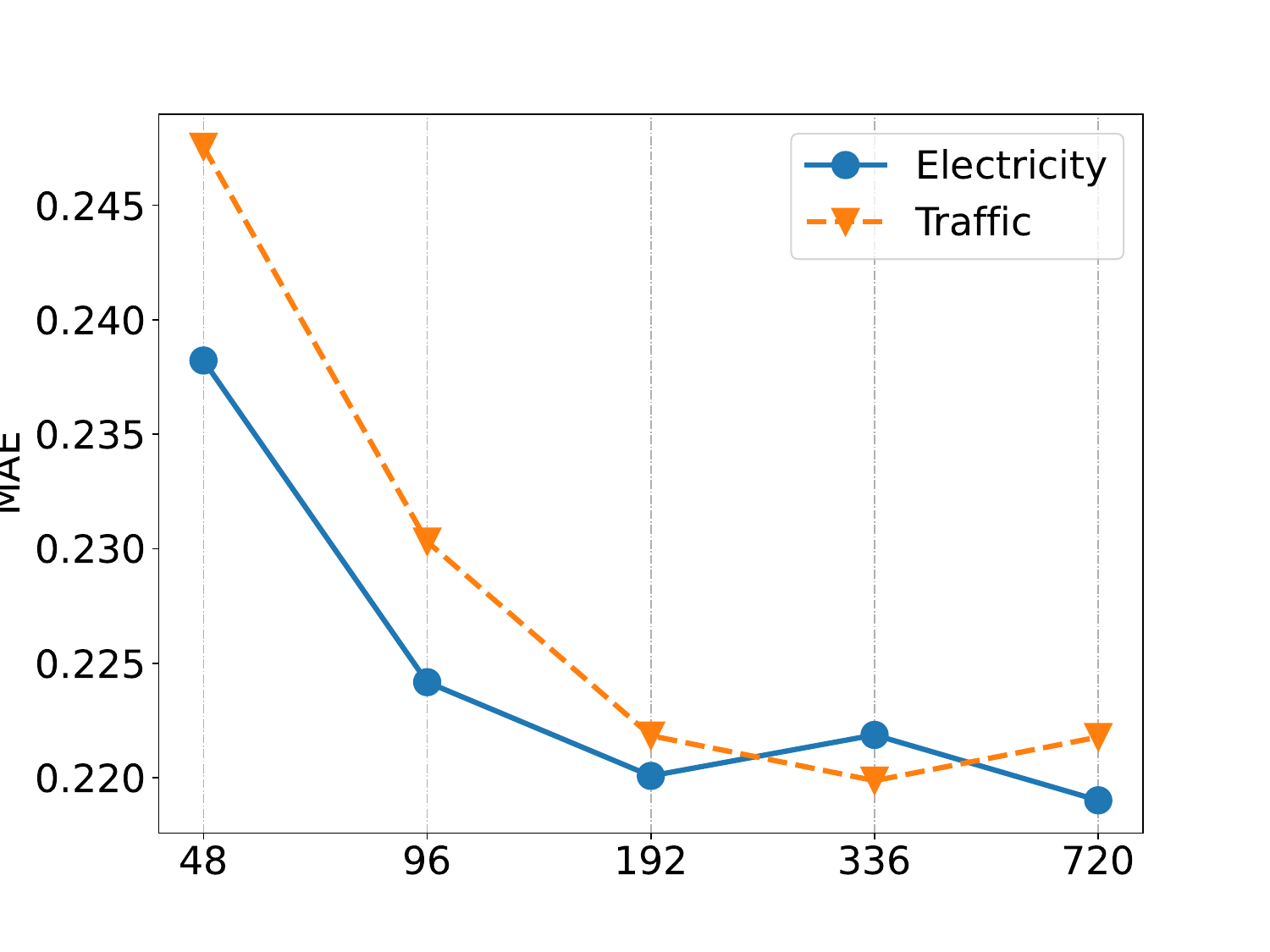}%
			\label{fig:lookback.2}}
		\caption{The MAE results (Y-axis) of models with different lookback window sizes (X-axis) of long-term forecasting (T=96) on the Traffic and Electricity datasets. (a) Transformer results; (b) PRformer results.}
		\label{fig:lookback}
	\end{minipage}
\end{figure}

The ability to use a longer review window reflects the model's ability to capture long-term dependency relationships. Many previous Transformer-based models have difficulty in utilizing longer review windows, as evidenced by unchanged or decreased predictive performance when the review window is increased, leading to doubts about the quality of Transformer models. PRformer uses multiscale pyramidal convolution and RNN to learn univariate time series representation. Since RNN is structurally aware of the sequence order of sequence data and has a high compatibility with time series data, and pyramidal convolution can turn long sequences into shorter sequences to alleviate the gradient vanishing problem of RNN, we speculate that PRformer has the ability to use longer review windows.The experimental results in the Figure \ref{fig:lookback} confirm this speculation: as the review window length increases, the prediction error of the ordinary Transformer model remains unchanged or increases, while the prediction error of PRformer shows a significant downward trend. This indicates that PRformer can utilize longer time series to improve overall predictive performance.

\begin{figure}[!t]
	\centering
	\vspace{0.1cm} 
	\begin{minipage}{1\columnwidth}
		\centering
		\subfloat[]{\includegraphics[width=0.45\linewidth]{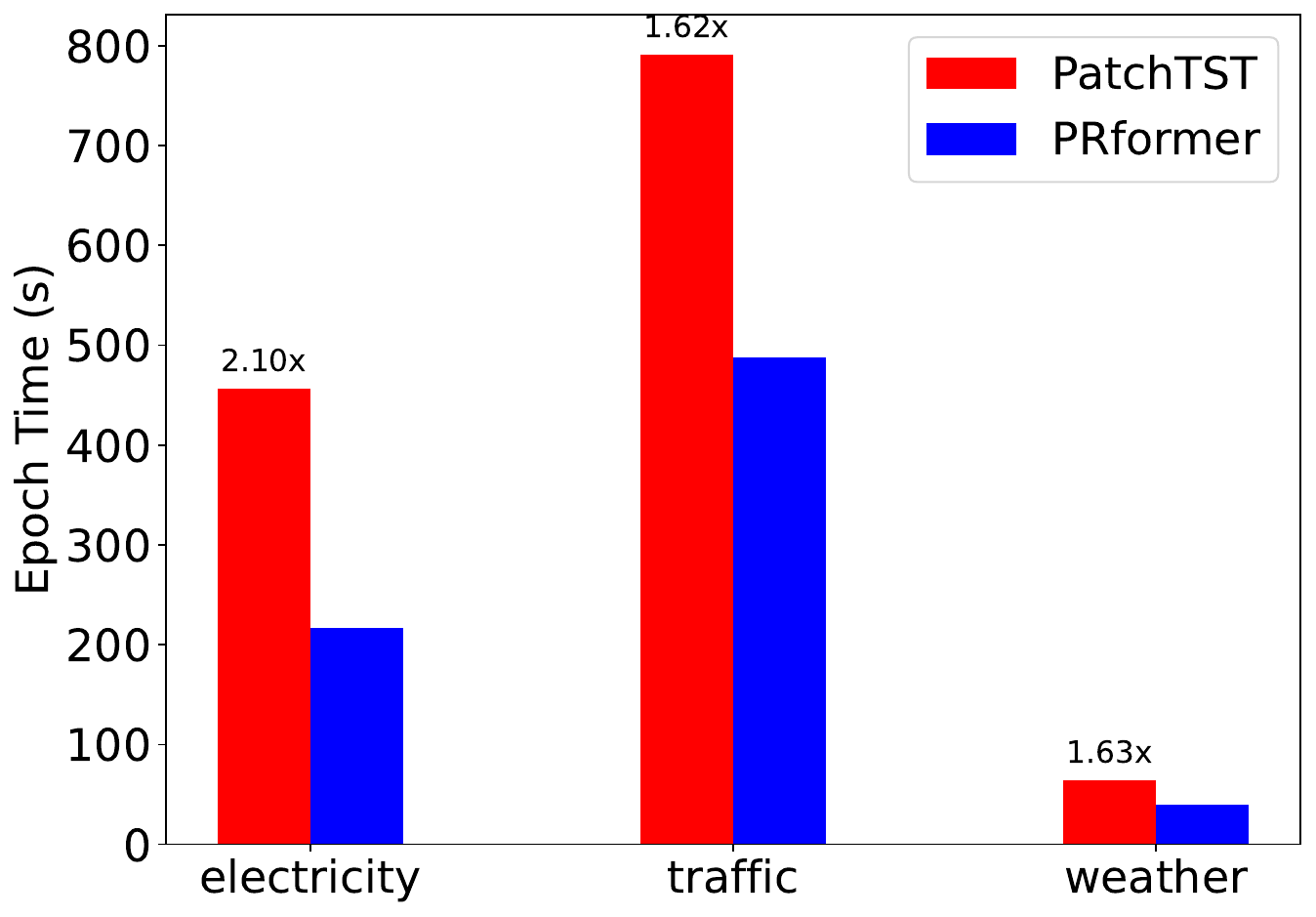}%
			\label{fig:epoch-time}}
		\hfill
		\subfloat[]{\includegraphics[width=0.45\linewidth]{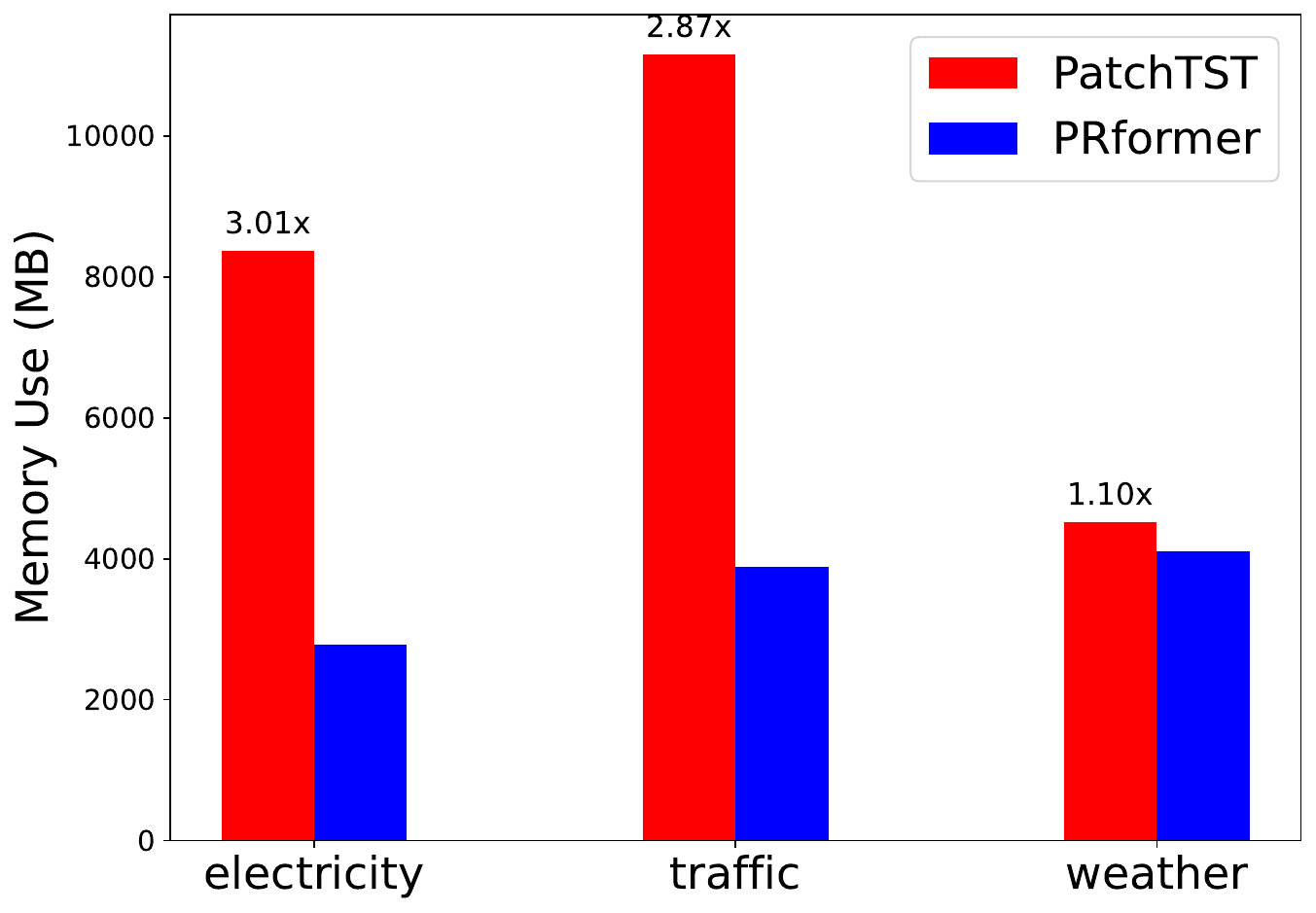}%
			\label{fig:memory-use}}
		\caption{In (a) we present the training times for one epoch on three datasets. In (b) we illustrate the corresponding memory usage. The experiments were conducted under equivalent hardware conditions and parameter configurations.}
		\label{fig:efficient}
	\end{minipage}
\end{figure}

\textbf{Model Efficiency}

To verify the actual efficiency of PRformer, we compared the memory usage and runtime of PatchTST and PRformer on the same GPU, maintaining identical lookback window length, prediction length, and batch size. The results from Figure \ref{fig:efficient} reveal that the runtime of PRformer is nearly half that of PatchTST, while its memory usage ranges between 1/3 and 0.9 of PatchTST's. This indicates that under equivalent hardware conditions, PRformer can accommodate a larger batch size, thereby further accelerating the training and inference speeds. Additionally, PRformer is capable of utilizing longer lookback windows, enabling it to learn dependencies over a larger range and thus further enhance its performance. This flexibility provides more opportunities for real-world applications and extensions.

\section{Conclusion}
In this paper, we explore a new direction for applying Transformers to time series prediction. We propose using a Pyramidal RNN Embedding (PRE) module to replace the positional encoding in Transformers, addressing the deficiencies of Transformers in encoding sequential positional information relationships. Through experiments combining PRE with three Transformer variants, we validate the significant improvement that PRE brings to Transformer architecture predictors. Furthermore, we propose PRformer, a time series predictor that combines PRE and Transformers. Comprehensive empirical experiments on eight long-term prediction benchmarks demonstrate the superiority of our approach. Additionally, PRE enables the computational complexity of Transformers to grow linearly with sequence length, resulting in better computational efficiency. We hope that this work will promote future research on the integration of RNNs and Transformers in time series modeling.

\bibliography{refs}

\appendix

\section{Limitations of Positional Encoding}

Transformer relies on positional encoding to determine the position of each timestamp(including relative and absolute distances), which may limit its capabilities and raise questions about its effectiveness compared to linear models \cite{zeng2023are}. The original positional encoding method in Transformers is as follows:

\begin{equation}
	\scalebox{1}{
		$\begin{aligned}
			PE_t^{(i)}&=
			\begin{cases}
				\sin\left(\frac{1}{10000^{2k/d_{model}}}t\right), & \text{if } i=2k\\
				\cos\left(\frac{1}{10000^{2k/d_{model}}}t\right), & \text{if } i=2k+1
			\end{cases}, \\
			&\quad i=0,1,2,3,\ldots,d_{model}-1
		\end{aligned}$
	}
\end{equation}
This design has the following property: the dot product of two positional encodings depends only on the offset \(\triangle t\) and is independent of the absolute positions.
Proof:

\begin{equation}
	\scalebox{0.85}{
		$\begin{aligned}
			&PE_{t}^{T} \cdot PE_{t+\triangle t} \\
			&= \sum_{i=0}^{\frac{d_{model}}{2}-1} [PE_t^{(2i)} \cdot PE_{t+\triangle t}^{(2i)} + PE_t^{(2i+1)} \cdot PE_{t+\triangle t}^{(2i+1)}] \\
			&= \sum_{i=0}^{\frac{d_{model}}{2}-1} [\sin(w_it) \cdot \sin(w_i(t+\triangle t)) + \cos(w_it) \cdot \cos(w_i(t+\triangle t))] \\
			&= \sum_{i=0}^{\frac{d_{model}}{2}-1} \cos(w_i\triangle t)
		\end{aligned}$
	}
\end{equation}

This property makes the distance measurement generated by positional encoding translation-invariant. When modeling time series data, traditional Transformers attempt to learn the variation patterns of sequences by combining multivariate data at different moments through positional encoding. However, real-world data is often non-stationary, and the patterns of data variation change significantly over time \cite{9373728}. Adopting the assumption of translation invariance makes it difficult for the self-attention mechanism of Transformers to learn the correct temporal dependencies. The DTW algorithm \cite{1163491} is a good example in this regard. By finding the optimal alignment path, DTW allows sequences to "stretch" to obtain more accurate sequence similarity. Due to the translation invariance of positional encoding, Transformers cannot dynamically adjust like the DTW algorithm to adapt to the similarity relationships of non-stationary sequences, thus limiting performance improvement.

Compared to Transformers, the recurrent structure of RNN does not have fixed positional encoding and can adaptively adjust to sequences of different lengths, which helps in handling non-stationary time series. In addition, RNN have strong function fitting capabilities \cite{graves2014neural,khrulkov2018expressive} and have the potential to transform non-stationary sequences into representations \cite{9127499} that are more conducive to Transformer learning. Based on these considerations, we combine RNN with PRE to transform non-stationary univariate time series into representations in vector space, and then use Transformers to learn the dependencies between variables. This avoids the limitations of traditional Transformers using positional encoding to learn the dependencies between points within sequences, resulting in a significant performance improvement.

\section{Experimental details}
\subsection{Dataset}
In our investigation, we evaluated the PRformer model using seven widely recognized datasets in the field of time series analysis. These datasets represent diverse domains and intervals, offering a comprehensive assessment environment:
The datasets used in our analysis can be categorized as follows:
\begin{itemize}
	
	\item \textbf{ETT}: Electricity Transformer Temperature (ETT) is a critical parameter in long-term electric power infrastructure planning. This dataset contains data spanning two years from two separate counties in China. It has been curated for use in long-sequence time-series forecasting (LSTF) research, and different subsets have been carefully constructed to explore the granularity of this problem. These subsets include ETTh1 and ETTh2, which provide data at 1-hour intervals, as well as ETTm1, which offers data at 15-minute intervals. 
	
	\item \textbf{Weather}: A comprehensive collection of 21 weather parameters, including temperature and humidity, meticulously logged at 10-minute intervals during the entire year of 2020.
	
	\item \textbf{Electricity}: This dataset chronicles the hourly electricity usage of 321 customers, covering a timespan from 2012 to 2014.
	
	\item \textbf{Traffic}: Sourced from the California Department of Transportation, this dataset offers insights into highway occupancy rates with data aggregated hourly from 862 sensors across the San Francisco Bay Area.
	
	\item \textbf{Solar-Energy}: This dataset consists of solar power production records from the year 2006, collected every 10 minutes from 137 photovoltaic (PV) plants in Alabama State. These data points provide comprehensive insights into the power output of hypothetical solar plants across the United States.

\end{itemize}

For our analysis, we divided these datasets into training, validation, and test segments in line with the protocols recommended in recent literature \cite{wu2021autoformer,zeng2023are,nie2023a}. Specifically, the ETT datasets were split in a 6:2:2 ratio, consistent with these guidelines. For the other datasets, a 7:1:2 division was employed. To ensure uniformity in our approach, we set the length of historical data sequences at 720 time steps, with the forecast horizon varying among the options of 96, 192, 336, and 720 time steps.

\subsection{Confiuration}

All experiments were implemented in PyTorch \cite{paszke2019pytorch} and conducted on a dedicated NVIDIA TITAN XP 11GB GPU. We employed the Adam optimizer \cite{kingma2014adam} for training the model over 30 epochs, with an initial learning rate decayed exponentially starting from the fourth epoch, using a decay factor of 0.9. The early stopping mechanism was configured with a patience value of 10.

\begin{table*}[htbp]
	\centering
	\caption{Parameter Configuration Table of PRformer Across Different Datasets.}
	
	\resizebox{0.9\textwidth}{!}{
		\begin{tabular}{lrlrrrrr}
			\toprule
			Datasets & \multicolumn{1}{l}{lookback} & pyramidal\_windows & \multicolumn{1}{l}{e\_layers} & \multicolumn{1}{l}{d\_model} & \multicolumn{1}{l}{dropout} & \multicolumn{1}{l}{batch\_size} & \multicolumn{1}{l}{lr} \\
			\midrule
			ETTh1 & 720   & 24 48 72 144 & 5     & 720   & 0.1   & 256   & 0.001 \\
			ETTh2 & 720   & 24 48 72 144 & 5     & 720   & 0.1   & 256   & 0.0002 \\
			ETTm1 & 720   & 4 16 32 96 & 5     & 720   & 0.1   & 256   & 0.0002 \\
			ETTm2 & 720   & 4 16 32 96 & 5     & 720   & 0.1   & 256   & 0.0001 \\
			Weather & 720   & 6 24 48 144 & 3     & 720   & 0.1   & 64    & 0.0001 \\
			Electricity & 720   & 24 48 72 96 144 & 3     & 660   & 0.1   & 16    & 0.0005 \\
			Traffic & 720   & 24 48 72 144 & 4     & 520   & 0.1   & 8     & 0.001 \\
			\bottomrule
		\end{tabular}%
	}
	\label{tab:parameter}%
\end{table*}%

The specific parameters utilized by PRformer on different datasets are outlined in Table \ref{tab:parameter}. The meaning of each parameter in the table is elucidated as follows:

\begin{itemize}
	\item \textbf{lookback}: Length of the historical lookback window.
	\item \textbf{pyramidal\_windows}: Sets the list of period lengths for the pyramidal convolution layers. The pyramidal convolution layers are generated based on this list.
	\item \textbf{e\_layers}: Number of Transformer encoder layers set.
	\item \textbf{d\_model}: The dimension of the embedding for univariate time series PRE.
	\item \textbf{dropout}: Dropout rate.
	\item \textbf{batch\_size}: Batch size used for training.
	\item \textbf{l rate (learning rate)}: The initial learning rate used in the optimization process.
\end{itemize}

\subsection{Implementation Details of PRE}

The pyramidal convolution structure in PRE is generated according to the set list of period lengths parameters \textit{pyramidal\_windows}. Each number in the \textit{pyramidal\_windows} list represents a period, corresponding to a convolutional recurrent neural network chain. The convolutional pyramid structure, composed of multiple chains, represents a period at each layer, with the period length increasing by multiples from the bottom to the top layer. Each chain generates a time series embedding representation for one period, and the embedding dimension of each chain is evenly distributed according to the number of pyramid layers, being $d\_model|layer\_num$. Finally, the results from all chains are weighted concatenated and transformed through a linear layer similar to a multi-head attention mechanism to obtain the PRE embedding representation. Both the input and output dimensions of the linear layer are $d\_model$.

\begin{table*}[htbp]
	\centering
	\caption{Full performance  promotion  results of Transformers  with PRE. }
	\resizebox{0.8\textwidth}{!}{
		\begin{tabular}{c|c|c|cc|cc|cc|cc}
			\toprule
			\multicolumn{3}{c|}{Models} & \multicolumn{2}{p{12.38em}|}{Transformer\newline{}(2017)} & \multicolumn{2}{p{12.25em}|}{Reformer\newline{}(2020)} & \multicolumn{2}{p{12.25em}|}{Informer\newline{}(2021)} & \multicolumn{2}{p{12.25em}}{Flowformer\newline{}(2022)} \\
			\cmidrule{4-11}    \multicolumn{3}{c|}{Metric} & MSE   & MAE   & MSE   & MAE   & MSE   & MAE   & MSE   & MAE \\
			\midrule
			\multirow{10}[8]{*}{Electricity} & \multirow{5}[4]{*}{Original} & 96    & 0.260  & 0.358  & 0.312  & 0.402  & 0.274  & 0.368  & 0.215  & 0.320  \\
			&       & 192   & 0.266  & 0.367  & 0.348  & 0.433  & 0.296  & 0.386  & 0.259  & 0.355  \\
			&       & 336   & 0.280  & 0.375  & 0.350  & 0.433  & 0.300  & 0.394  & 0.296  & 0.383  \\
			&       & 720   & 0.302  & 0.386  & 0.340  & 0.420  & 0.373  & 0.439  & 0.296  & 0.380  \\
			\cmidrule{3-11}          &       & Avg   & 0.277  & 0.372  & 0.338  & 0.422  & 0.311  & 0.397  & 0.267  & 0.360  \\
			\cmidrule{2-11}          & \multirow{5}[4]{*}{\textbf{+PRE}} & 96    & 0.128  & 0.218  & 0.131  & 0.227  & 0.130  & 0.226  & 0.132  & 0.228  \\
			&       & 192   & 0.148  & 0.238  & 0.149  & 0.244  & 0.149  & 0.243  & 0.150  & 0.244  \\
			&       & 336   & 0.162  & 0.254  & 0.166  & 0.262  & 0.166  & 0.262  & 0.166  & 0.261  \\
			&       & 720   & 0.184  & 0.276  & 0.206  & 0.297  & 0.203  & 0.295  & 0.197  & 0.289  \\
			\cmidrule{3-11}          &       & Avg   & \textbf{0.156 } & \textbf{0.247 } & \textbf{0.163 } & \textbf{0.258 } & \textbf{0.162 } & \textbf{0.257 } & \textbf{0.161 } & \textbf{0.256 } \\
			\midrule
			\multirow{10}[8]{*}{Traffic} & \multirow{5}[4]{*}{Original} & 96    & 0.647  & 0.357  & 0.732  & 0.423  & 0.719  & 0.391  & 0.691  & 0.393  \\
			&       & 192   & 0.649  & 0.356  & 0.733  & 0.420  & 0.696  & 0.379  & 0.729  & 0.419  \\
			&       & 336   & 0.667  & 0.364  & 0.742  & 0.420  & 0.777  & 0.420  & 0.756  & 0.423  \\
			&       & 720   & 0.697  & 0.376  & 0.755  & 0.432  & 0.864  & 0.472  & 0.825  & 0.449  \\
			\cmidrule{3-11}          &       & Avg   & 0.665  & 0.363  & 0.741  & 0.424  & 0.764  & 0.416  & 0.750  & 0.421  \\
			\cmidrule{2-11}          & \multirow{5}[4]{*}{\textbf{+PRE}} & 96    & 0.353  & 0.222  & 0.362  & 0.257  & 0.363  & 0.257  & 0.358  & 0.256  \\
			&       & 192   & 0.372  & 0.233  & 0.381  & 0.265  & 0.380  & 0.265  & 0.378  & 0.265  \\
			&       & 336   & 0.385  & 0.241  & 0.397  & 0.274  & 0.396  & 0.273  & 0.395  & 0.273  \\
			&       & 720   & 0.421  & 0.258  & 0.436  & 0.297  & 0.436  & 0.296  & 0.435  & 0.296  \\
			\cmidrule{3-11}          &       & Avg   & \textbf{0.383 } & \textbf{0.239 } & \textbf{0.394 } & \textbf{0.273 } & \textbf{0.394 } & \textbf{0.273 } & \textbf{0.392 } & \textbf{0.272 } \\
			\midrule
			\multirow{10}[8]{*}{Weather} & \multirow{5}[4]{*}{Original} & 96    & 0.395  & 0.427  & 0.689  & 0.596  & 0.300  & 0.384  & 0.182  & 0.233  \\
			&       & 192   & 0.619  & 0.560  & 0.752  & 0.638  & 0.598  & 0.544  & 0.250  & 0.288  \\
			&       & 336   & 0.689  & 0.594  & 0.639  & 0.596  & 0.578  & 0.523  & 0.309  & 0.329  \\
			&       & 720   & 0.926  & 0.710  & 1.130  & 0.792  & 1.059  & 0.741  & 0.404  & 0.385  \\
			\cmidrule{3-11}          &       & Avg   & 0.657  & 0.573  & 0.803  & 0.656  & 0.634  & 0.548  & 0.286  & 0.309  \\
			\cmidrule{2-11}          & \multirow{5}[4]{*}{\textbf{+PRE}} & 96    & 0.144  & 0.187  & 0.147  & 0.198  & 0.146  & 0.200  & 0.149  & 0.203  \\
			&       & 192   & 0.188  & 0.233  & 0.191  & 0.241  & 0.197  & 0.245  & 0.199  & 0.249  \\
			&       & 336   & 0.241  & 0.274  & 0.243  & 0.283  & 0.256  & 0.291  & 0.243  & 0.284  \\
			&       & 720   & 0.326  & 0.332  & 0.314  & 0.334  & 0.324  & 0.342  & 0.316  & 0.337  \\
			\cmidrule{3-11}          &       & Avg   & \textbf{0.225 } & \textbf{0.257 } & \textbf{0.224 } & \textbf{0.264 } & \textbf{0.231 } & \textbf{0.269 } & \textbf{0.227 } & \textbf{0.268 } \\
			\bottomrule
		\end{tabular}%
	}
	\label{tab:full_promotion}%
\end{table*}%

\section{Full PRE promotion results}
To validate the effectiveness and generalizability of the PRE in enhancing the Transformer's ability to learn temporal sequence representations, we applied PRE to the Transformer and its variants: Transformer\cite{vaswani2017attention}, Reformer \cite{kitaev2020reformer}, Informer\cite{zhou2021informer},and  Flowformer \cite{wu2022flowformer}. Due to space constraints, Table \ref{tab:promotion_avg} presents the average results, while the complete predictive outcomes are detailed in Table \ref{tab:full_promotion}. The results demonstrate that PRE consistently improves the performance of these Transformers, thereby affirming its efficacy and generalization capabilities.

\section{Univariate Time Series Representation Learning}

\begin{figure*}[!t]
	\centering
	\subfloat[ETTh1s]{\includegraphics[width=0.2\textwidth]{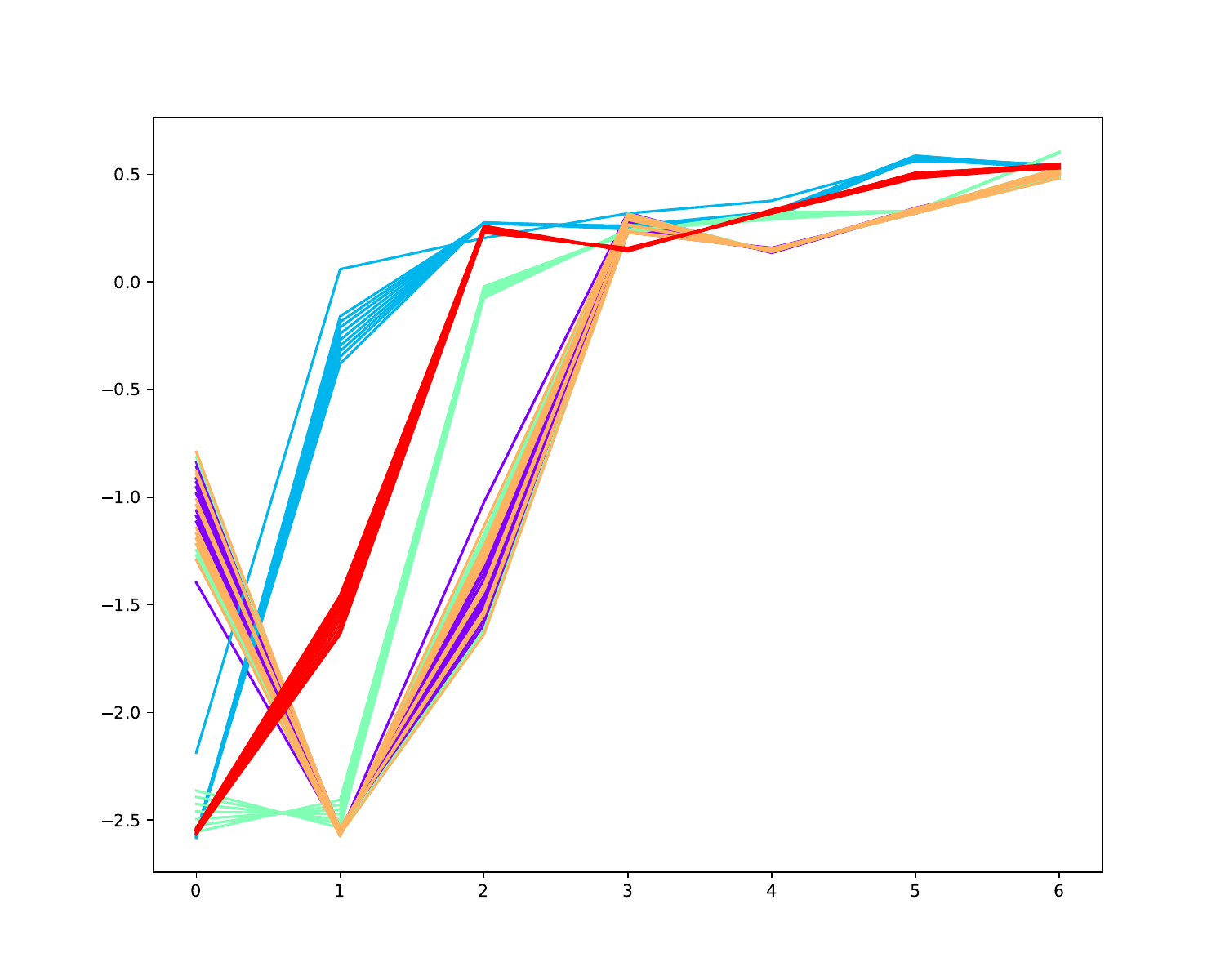}%
		\label{fig:pre_origin.1}}
	\hfil
	\subfloat[ETTh2]{\includegraphics[width=0.2\textwidth]{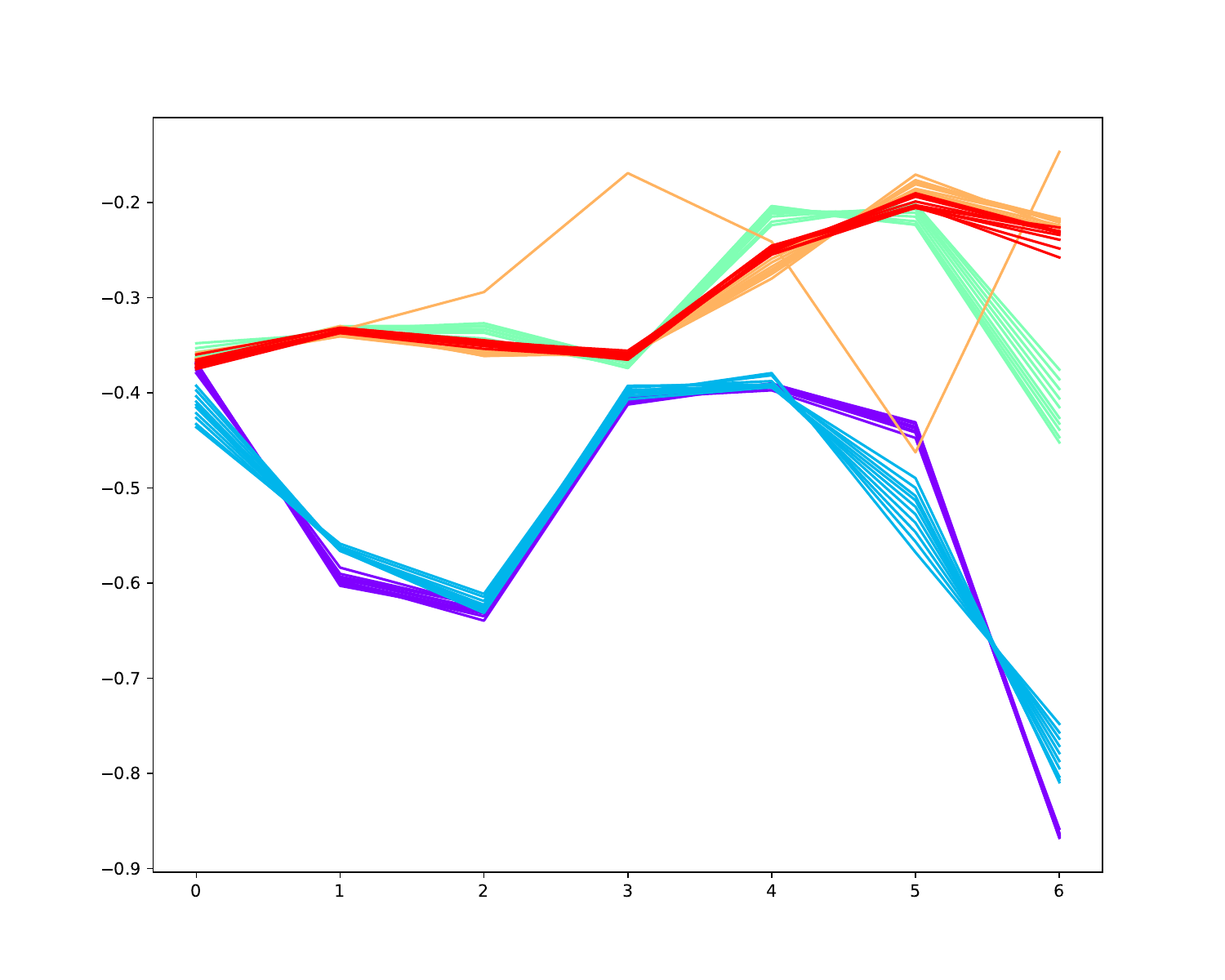}%
		\label{fig:pre_origin.2}}
	\hfil
	\subfloat[ETTm1]{\includegraphics[width=0.2\textwidth]{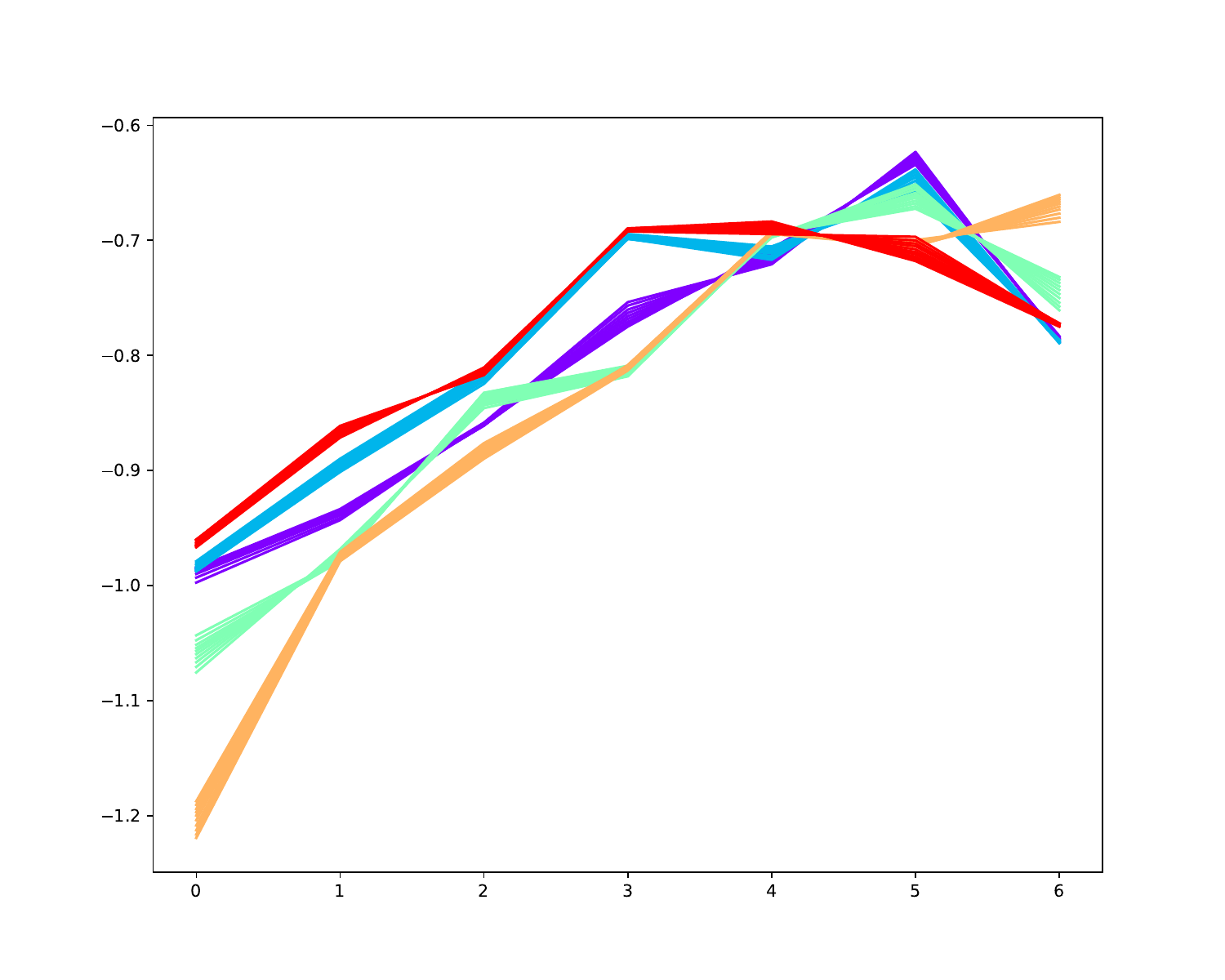}%
		\label{fig:pre_origin.3}}
	\hfil
	\subfloat[ETTm2]{\includegraphics[width=0.2\textwidth]{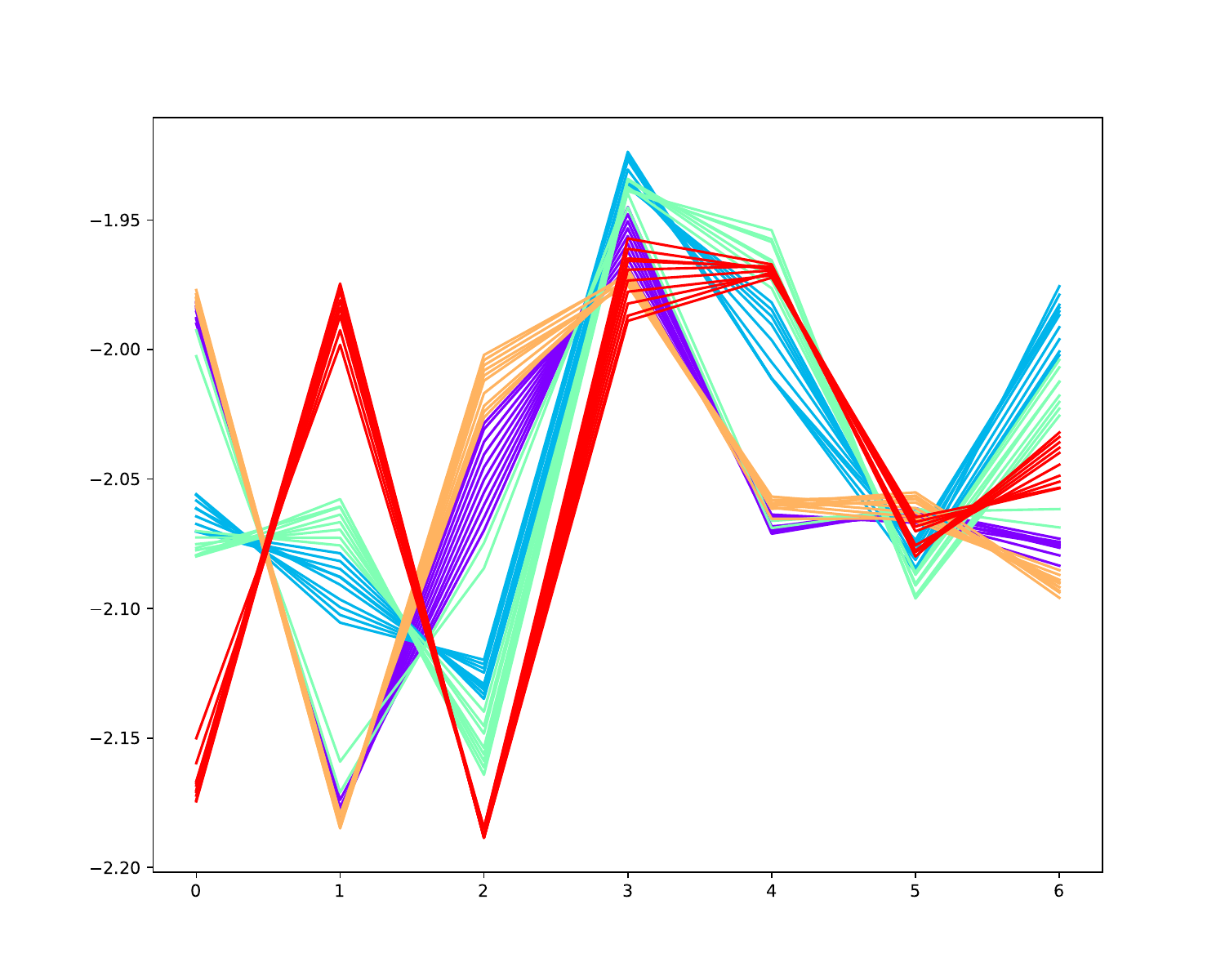}%
		\label{fig:pre_origin.4}}
	\hfil
	\subfloat[Weather]{\includegraphics[width=0.2\textwidth]{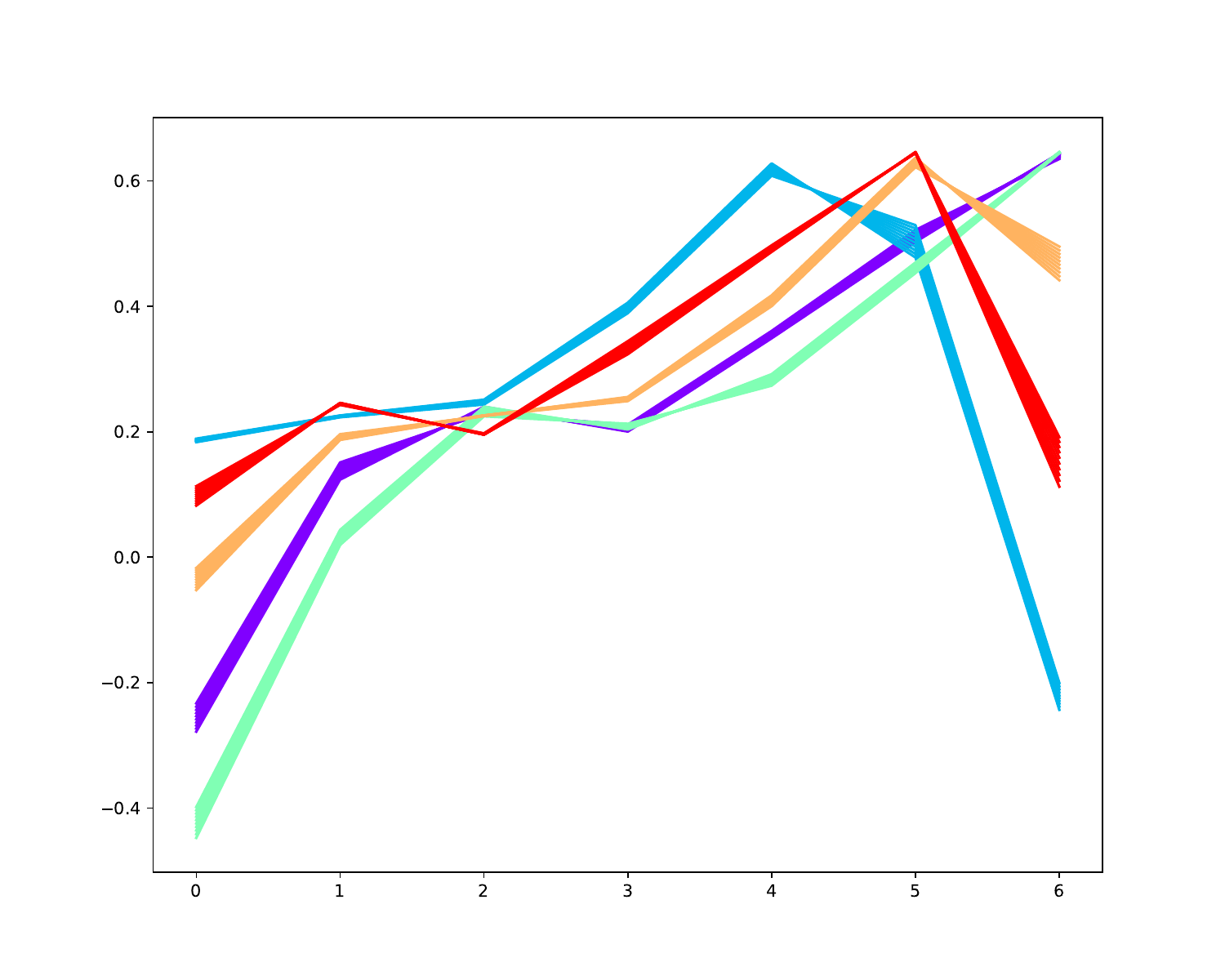}%
		\label{fig:pre_origin.5}}
	\hfil
	
	\subfloat[ETTh1]{\includegraphics[width=0.2\textwidth]{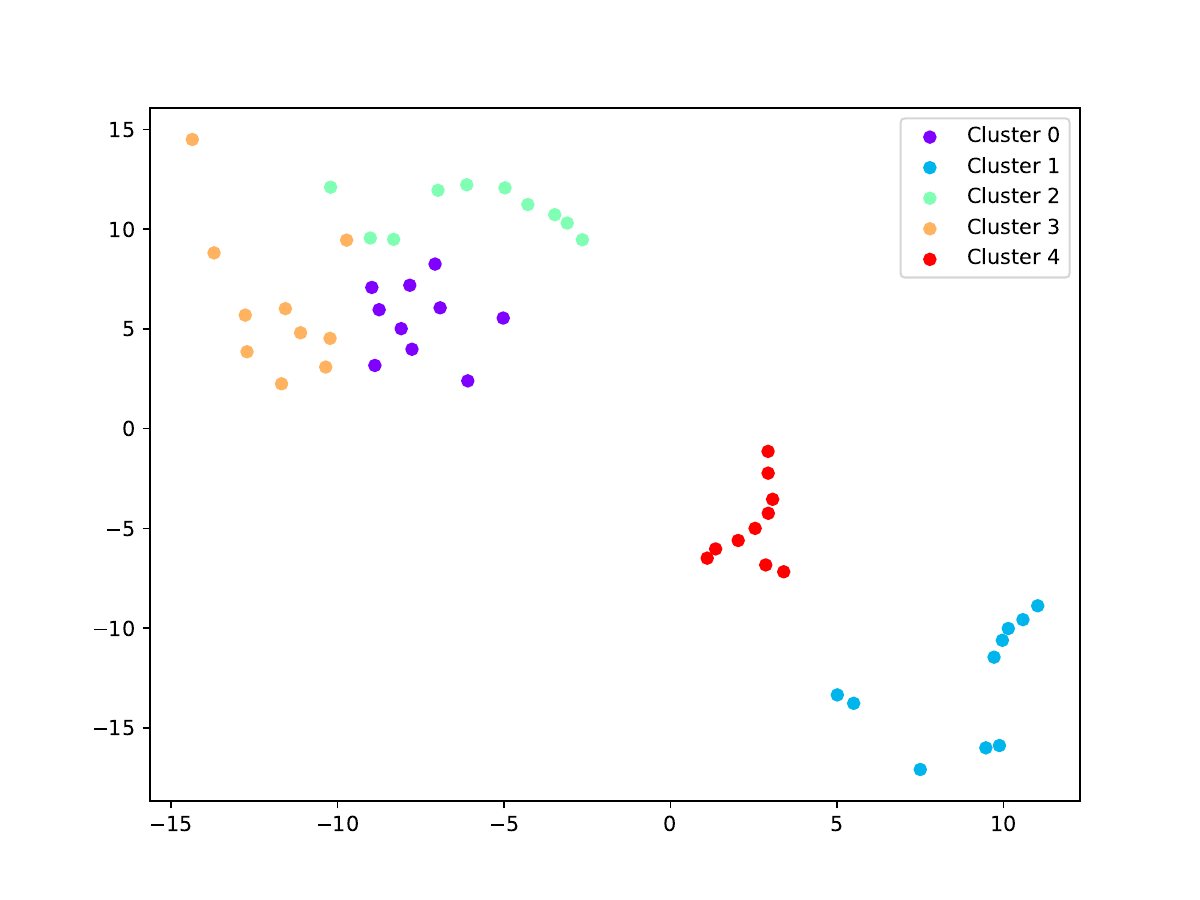}%
		\label{fig:pre_cluster.1}}
	\hfil
	\subfloat[ETTh2]{\includegraphics[width=0.2\textwidth]{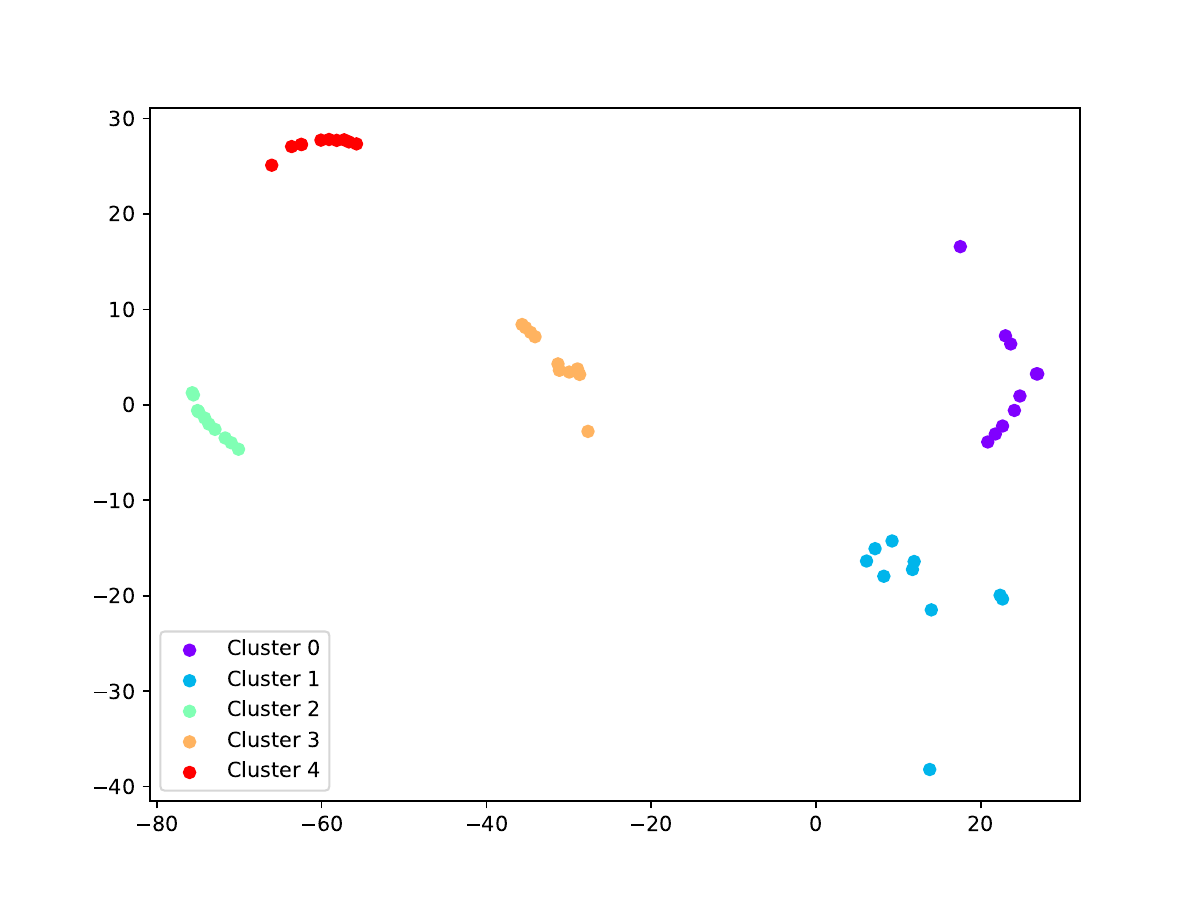}%
		\label{fig:pre_cluster.2}}
	\hfil
	\subfloat[ETTm1]{\includegraphics[width=0.2\textwidth]{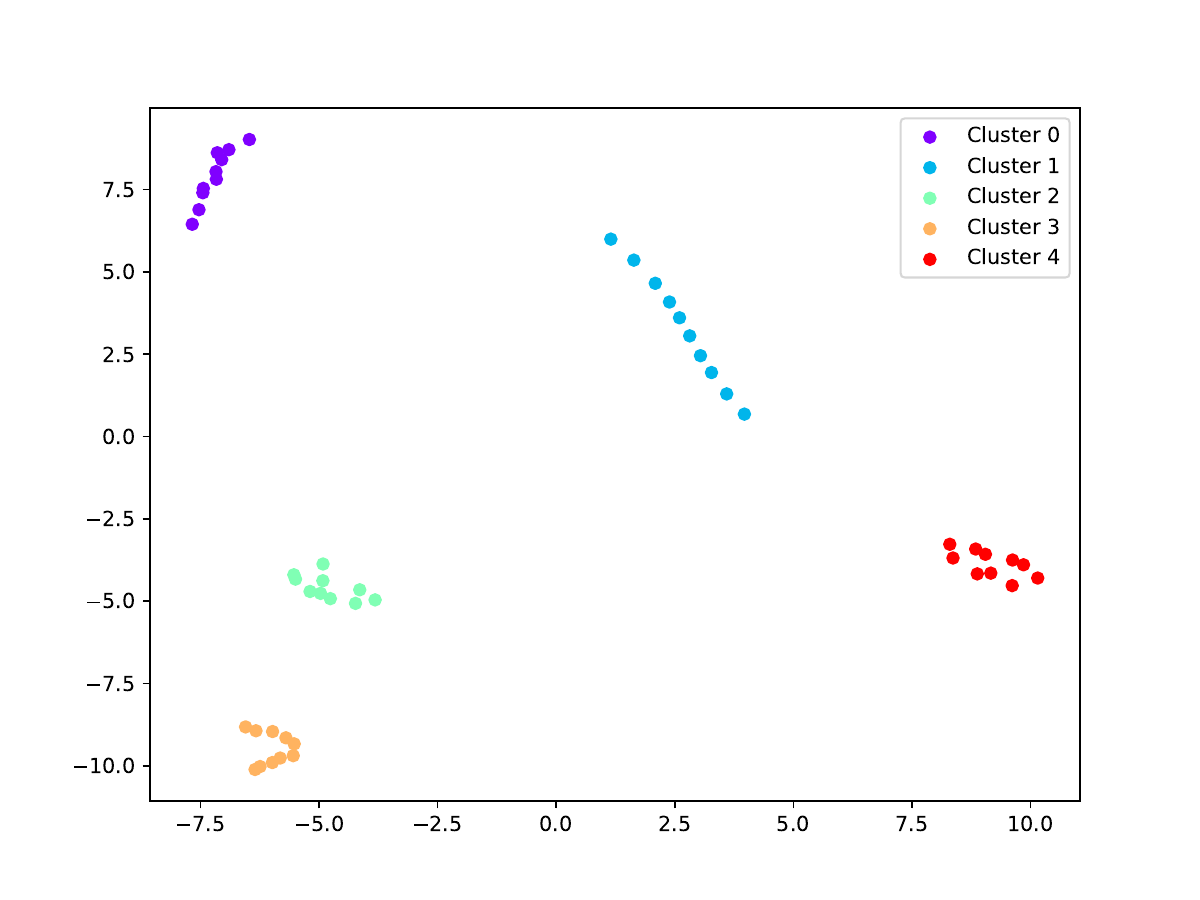}%
		\label{fig:pre_cluster.3}}
	\hfil
	\subfloat[ETTm2]{\includegraphics[width=0.2\textwidth]{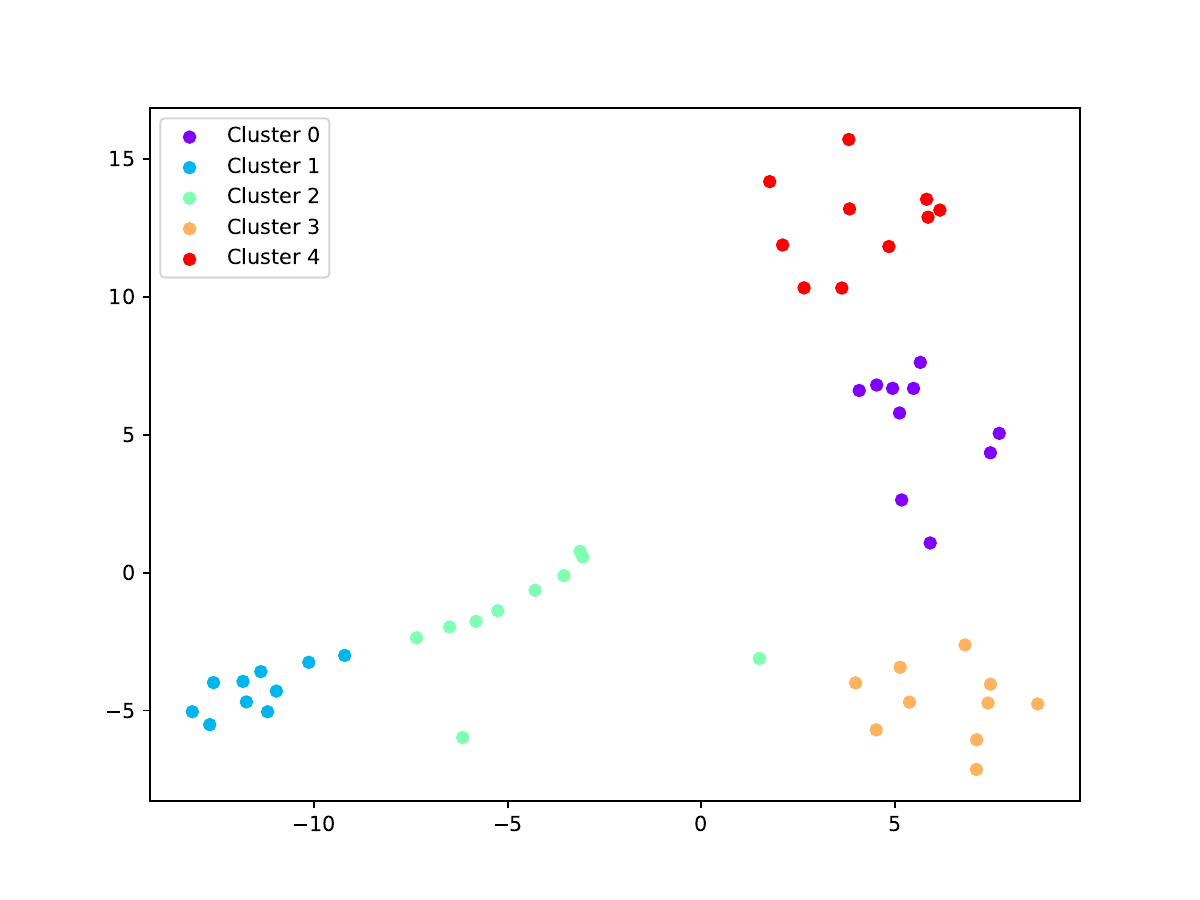}%
		\label{fig:pre_cluster.4}}
	\hfil
	\subfloat[Weather]{\includegraphics[width=0.2\textwidth]{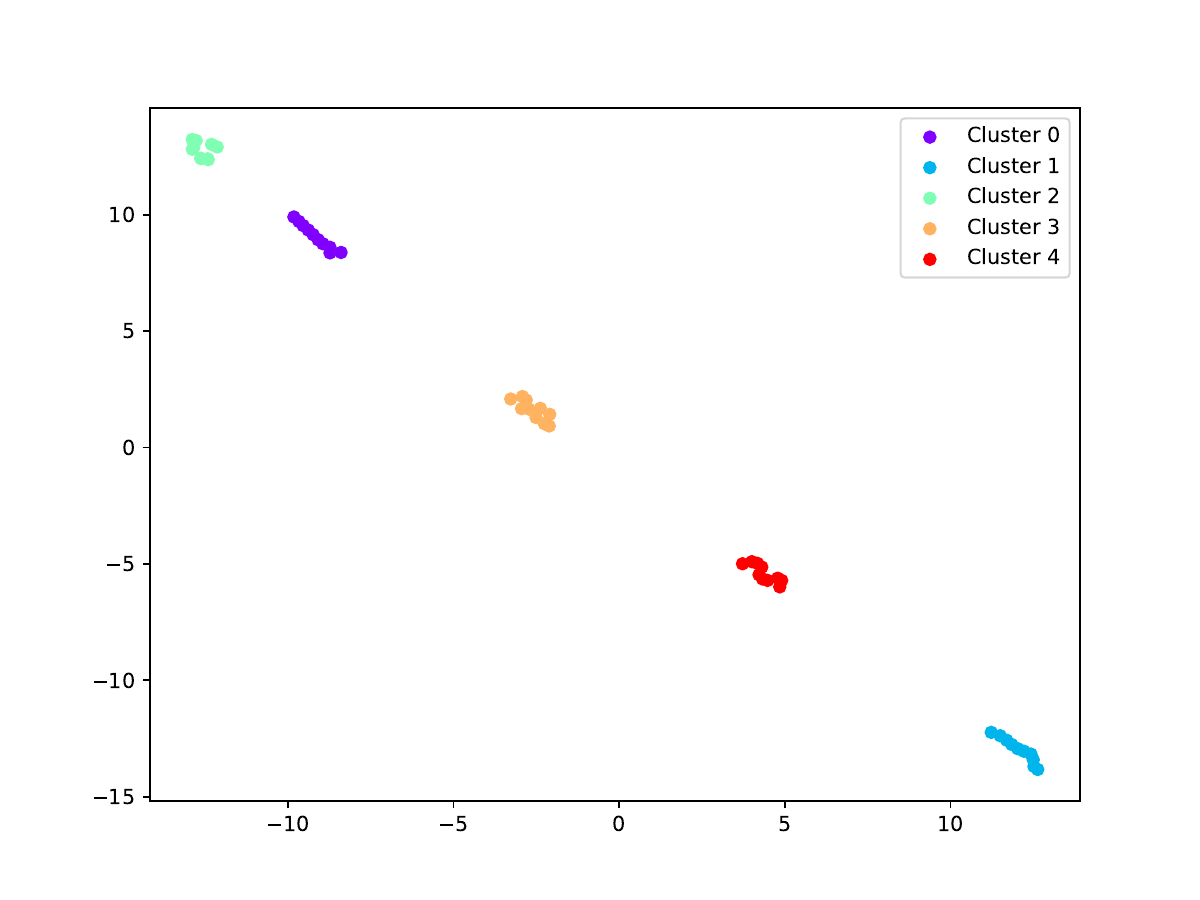}%
		\label{fig:pre_cluster.5}}
	\hfil
	\caption{Correlation of origin time series with their respective PRE embeddings across multiple datasets. Figures a-e represent the original time series, while figures f-j depict the corresponding t-SNE clustering results of PRE embeddings.Similar representations correspond to original sequences with analogous shapes and patterns, whereas distant representations showcase more substantial deviations in shape, thereby acquiring meaningful time series representations.}
	\label{fig:all_series_pre}
\end{figure*}

PRE provides a novel approach for time series prediction with Transformer. Initially, single-variable time series are represented using PRE. Subsequently, the Transformer attention mechanism is employed to learn intricate dependencies among multiple variables and make predictions. To analyze PRE's learned representations, 50 embeddings were randomly chosen from the test set of multiple datasets, post-PRE training. We then applied t-SNE\cite{van2008visualizing} for dimensionality reduction, forming five distinct clusters in the embedding space. This is visualized in Figure \ref{fig:all_series_pre}.From the figures, we observe that similar representations correspond to original sequences with similar shapes and patterns, while distant representations exhibit larger shape differences. This observation underscores PRE's ability to generate meaningful embeddings for single variables, aiding the Transformer in constructing dependencies among multiple variables and subsequent prediction tasks.

\end{document}